\documentclass[letterpaper]{article} 
\usepackage[preprint]{aaai2027}  
\usepackage[hyphens]{url}  
\usepackage{graphicx} 
\usepackage{natbib}  
\usepackage{caption} 
\usepackage{booktabs}
\usepackage{colortbl}
\newcolumntype{L}[1]{>{\raggedright\arraybackslash}p{#1}}
\newcolumntype{C}[1]{>{\centering\arraybackslash}p{#1}}
\newcommand{\projectpagelink}{%
  \leavevmode
  \pdfstartlink attr{/Border [0 0 0]} user{%
    /Subtype /Link /A << /S /URI /URI (https://physmind.github.io/) >>}%
  \url{https://physmind.github.io/}%
  \pdfendlink}

\title{PhysMind: From Video to Executable Worlds\\
for Training-Free Physical Reasoning}
\author{
    Chen Yang\textsuperscript{\rm 1}\equalcontrib,
    Shenxiang Zeng\textsuperscript{\rm 1}\equalcontrib,
    Haoyang Zhao\textsuperscript{\rm 1},
    Zhouyuan Xu\textsuperscript{\rm 1},
    Youquan He\textsuperscript{\rm 1},\\
    Haoyu Li\textsuperscript{\rm 1},
    Mingyi Deng\textsuperscript{\rm 2},
    Jiansheng Fan\textsuperscript{\rm 1},
    Chen Wang\textsuperscript{\rm 1}\corresponding
}
\affiliations{
    \textsuperscript{\rm 1}Tsinghua University\\
    \textsuperscript{\rm 2}The University of Hong Kong\\
    chwang@tsinghua.edu.cn
}

\begin{document}

\maketitle

\begin{abstract}
Reliable physical reasoning from video requires understanding how objects move, interact, and respond to interventions. Existing vision-language models (VLMs) often struggle to interpret these dynamics and reason reliably about future and counterfactual outcomes. We introduce PhysMind, a training-free agentic framework that constructs one reusable, question-agnostic executable world per video. PhysMind recovers a temporally consistent dynamic scene through object segmentation, mesh reconstruction, and 6D pose tracking, then fits analytic continuous-time dynamics and latent physical parameters without unrolling a time-stepped simulator. Given a question, it inspects, continues, or edits the world and answers from the resulting trajectories and interactions. Relative to direct chain-of-thought (CoT) reasoning with the same VLM, PhysMind improves accuracy by 38.23 points on CLEVRER and 8.08 points on Physion++. On counterfactual questions, it exceeds the strongest evaluated VLM baseline, GPT-5.5, by 19.25 points. Project page: \projectpagelink.
\end{abstract}


\section{Introduction}

Vision-language models (VLMs) now exhibit strong image, video, and language understanding \cite{comanici2025gemini25,bai2025qwen3vl}. Their ability to reason about physical processes in video, however, remains limited \cite{krojer2025mvp}, especially for explaining events, predicting outcomes, and assessing interventions \cite{balazadeh2025pcbs,tang2026causalphys}. Recent evaluations reveal unreliable reasoning about stability, collisions, latent material properties \cite{chow2025physbench}, object permanence \cite{bordes2025intphys2}, and quantitative kinematics \cite{puyin2026quantiphy}. Such weaknesses restrict embodied decision-making in robotic manipulation \cite{zhou2025physvlm} and autonomous driving \cite{lu2025realad}, where actions depend on how the world evolves rather than only how it appears.

Improving video-based physical reasoning is a long-standing challenge \cite{yi2020clevrer,tung2023physionpp}. Earlier approaches trained task-specific models to recover trajectories, learn interaction patterns, or predict future states \cite{chen2021dcl,ding2021dynamic,li2025slotpi}. Recent work instead augments general-purpose VLMs beyond direct CoT prompting. One direction uses supervised fine-tuning or reinforcement learning to induce physically grounded descriptions and reasoning traces \cite{balazadeh2025pcbs,wang2025videorft,feng2025videor1,wu2026videothinker,ghazanfari2025chainofframes}; another supplies inference-time evidence through frame memories, perception tools, or simulators \cite{chow2025physbench,fan2024videoagent,fan2025star,cherian2026llmphy}.

\begin{figure*}[t]
\centering
\includegraphics[width=\textwidth]{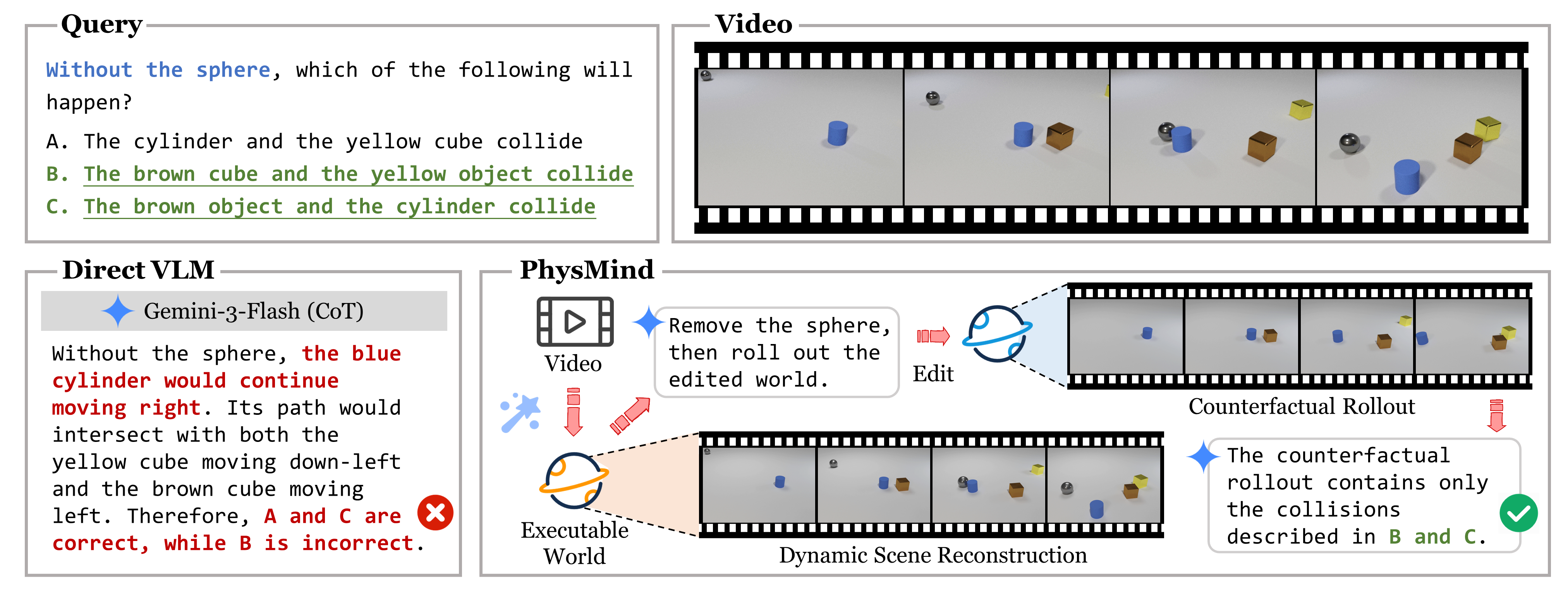}
\caption{PhysMind converts video into a reusable executable world. In this example, direct CoT reasoning predicts unsupported collisions after the sphere is removed. PhysMind instead edits the reconstructed world, executes the counterfactual rollout, and grounds its answer in the resulting collision events.}
\label{fig:teaser}
\end{figure*}

Despite this progress, most methods do not expose an explicit state-transition model of the observed scene. Predictions are derived from latent features, textual reasoning, or query-specific evidence, leaving the physical states and mechanisms connecting observations to answers unavailable for inspection or execution \cite{krojer2025mvp,tang2026causalphys}. Training-based approaches further require data and computation for adaptation and can trade off general VLM capabilities \cite{balazadeh2025pcbs,huang2025downstream}. This motivates a training-free approach that grounds VLM reasoning in an explicit, executable model recovered from video.

A natural route is to convert video into a physical world and answer questions by executing it. Advances in single-view and multi-view reconstruction, generation, and dynamic-scene recovery, including VGGT~\cite{wang2025vggt}, SAM 3D Objects~\cite{chen2026sam3d}, TRELLIS~\cite{xiang2025trellis}, and MonST3R~\cite{zhang2025monst3r}, provide the geometric foundation. Geometry alone does not determine how a scene evolves; latent mass, friction, restitution, and initial motion must also be estimated. System identification offers the complementary machinery for recovering these properties \cite{ding2021dynamic,cherian2026llmphy}.

However, enabling a VLM to coordinate perception, reconstruction, tracking, and dynamics estimation into one consistent executable world remains challenging. To this end, we propose PhysMind, a training-free agentic framework that orchestrates these tools and fits analytic dynamics to the recovered motion. As illustrated in Figure~\ref{fig:teaser}, PhysMind constructs the world once per video, reuses it across questions, and decides whether to inspect, continue, or edit its dynamics. The resulting trajectories, parameters, and outcomes provide inspectable evidence for open- and closed-source VLMs. PhysMind attains the highest overall accuracy among the evaluated methods on CLEVRER and Physion++.

Our contributions are threefold. First, we introduce PhysMind, a training-free agentic framework that turns video into a reusable executable world and grounds answers in its outcomes. Second, we develop two core components: dynamic scene reconstruction for persistent identities, conditioned meshes, and 6D trajectories; and analytic continuous-time system identification for efficient long-horizon dynamics fitting. Third, we evaluate PhysMind across six scenario families and four task types on CLEVRER and Physion++, improving direct CoT by 38.23 points on CLEVRER and exceeding GPT-5.5 by 19.25 points on counterfactual reasoning.

\begin{figure*}[t]
\centering
\includegraphics[width=\textwidth]{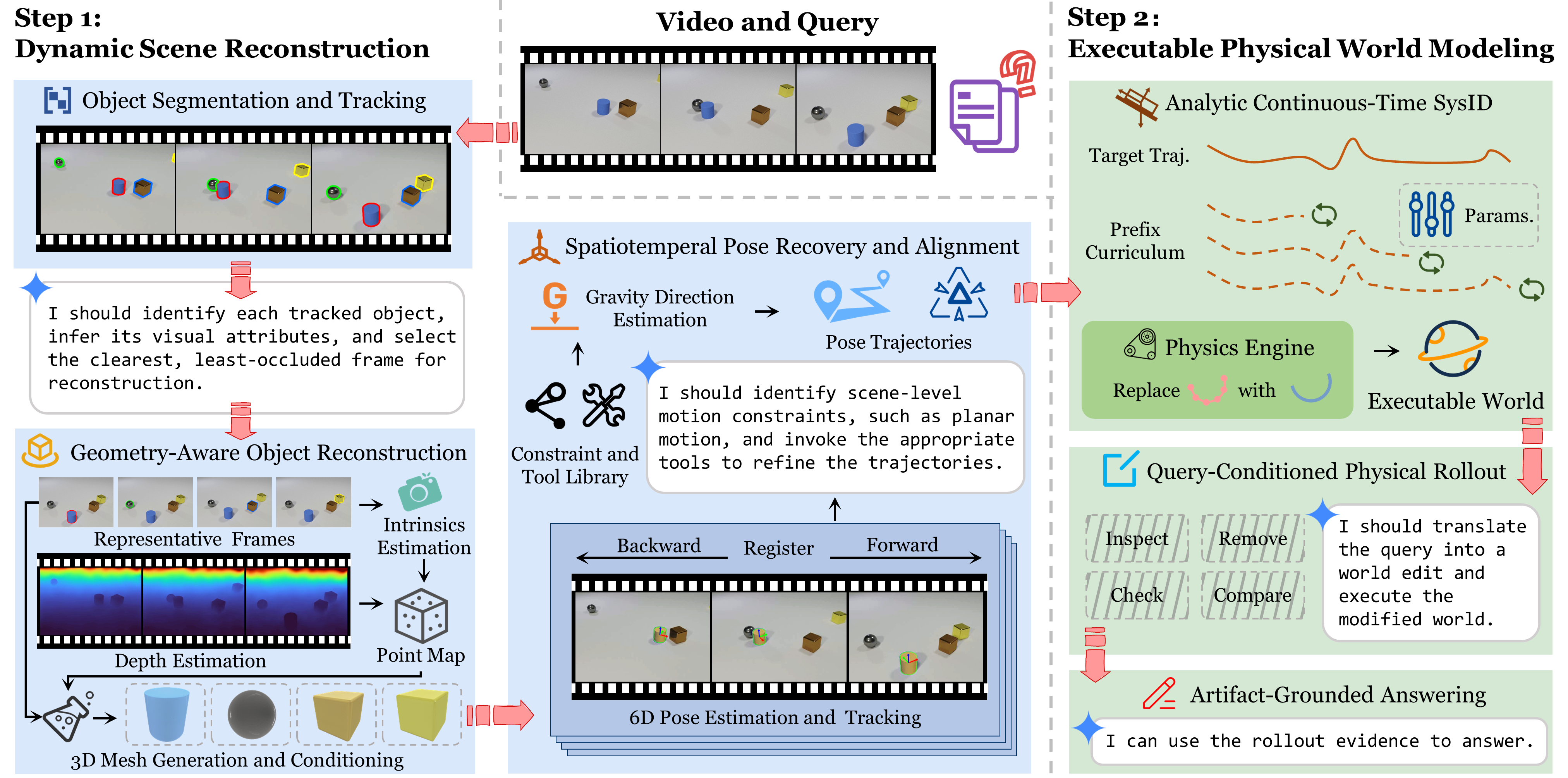}
\caption{Overview of PhysMind. Dynamic scene reconstruction recovers object tracks, conditioned meshes, and aligned 6D pose trajectories, with the VLM selecting scene-dependent constraints and tools. Executable physical world modeling fits analytic continuous-time dynamics, supports query-conditioned inspection and intervention, and grounds the final answer in execution artifacts. The question is used only after the reusable scene-level world has been constructed.}
\label{fig:overview}
\end{figure*}

\section{Related Work}

\paragraph{VLMs for Physical Reasoning} Task-specific approaches such as DCL~\cite{chen2021dcl}, VRDP~\cite{ding2021dynamic}, and SlotPi~\cite{li2025slotpi} learn object dynamics for individual benchmarks. Recent VLM methods follow two strategies. PCBs~\cite{balazadeh2025pcbs}, VideoRFT~\cite{wang2025videorft}, Video-R1~\cite{feng2025videor1}, VideoThinker-R1~\cite{wu2026videothinker}, and Chain-of-Frames~\cite{ghazanfari2025chainofframes} use targeted training to improve physical descriptions, temporal evidence selection, or structured reasoning. Training-free systems instead augment inference: PhysAgent~\cite{chow2025physbench} invokes perception modules, VideoAgent~\cite{fan2024videoagent} maintains frame memory, STAR~\cite{fan2025star} composes spatiotemporal tools, and LLMPhy~\cite{cherian2026llmphy} searches simulator parameters. These methods mainly infer from learned patterns or query-specific evidence. PhysMind instead builds one question-agnostic executable world, making physical reasoning explicit and reusable across tasks.

\paragraph{Agentic 3D Reconstruction} Agentic reconstruction coordinates pretrained vision models, 3D generators, optimization, and feedback to construct explicit scenes at inference time. CAST~\cite{yao2025cast} and VIGA~\cite{yin2026viga} decompose scenes and iteratively align object hypotheses. pySpatial~\cite{luo2026pyspatial} and GCA~\cite{chen2026gca} use reconstruction as a geometric workspace for spatial reasoning, while SIMPACT~\cite{liu2026simpact} and EToT~\cite{xu2025etot} construct interactive environments for manipulation planning. Closest to video-to-simulation, Vid2Sim~\cite{chen2025vid2sim} and PhysSplat~\cite{zhao2025physsplat} recover dynamic scene representations for downstream physical interaction. PhysMind differs in reconstructing persistent object geometry and 6D motion specifically as observations for system identification, coupling dynamic reconstruction to an executable physical model.

\paragraph{Physical System Identification} Contact-aware identification must accommodate nonsmooth changes in contact mode. Dojo~\cite{howell2022dojo} differentiates through hard-contact simulation, while Bianchini et al.~\cite{bianchini2023simultaneous} jointly learn contact and continuous dynamics. These methods optimize within a specified physical system. From visual trajectories, VRDP~\cite{ding2021dynamic} differentiates through an impulse-based simulator, whereas LLMPhy~\cite{cherian2026llmphy} uses trajectory error to guide black-box parameter search in MuJoCo or PyBullet. Time-stepped differentiation becomes costly and can yield weak gradients over long horizons; black-box search avoids this gradient path but still evaluates discrete rollouts. PhysMind instead fits continuous-time analytic trajectories and represents collisions as instantaneous state transitions, shortening the optimization path while preserving discontinuous interactions.

\section{Methodology}

PhysMind proceeds through two sequential stages: dynamic scene reconstruction recovers temporally consistent geometry and motion, after which executable physical world modeling fits and executes the recovered dynamics. As shown in Figure~\ref{fig:overview}, a VLM interprets scene-dependent conditions and selects specialized tools during world construction; once a question is provided, it composes query-conditioned operations and answers from the resulting execution artifacts.

\subsection{Problem Formulation}

Given a video $\mathcal{V}=\{I_t\}_{t=1}^{T}$, a question $q$, and an answer space $\mathcal{A}(q)$, PhysMind predicts $\hat{a}\in\mathcal{A}(q)$ through an explicit world model:
\begin{equation}
\begin{array}{rcl}
\mathcal{S} &=& F_{\mathrm{scene}}(\mathcal{V}),\\
\mathcal{W} &=& F_{\mathrm{id}}(\mathcal{S}),\\
\mathcal{R}_{q} &=& F_{\mathrm{exec}}(\mathcal{W},q),\\
\hat{a} &=& F_{\mathrm{ans}}(\mathcal{V},q,\mathcal{W},\mathcal{R}_{q}).
\end{array}
\label{eq:framework}
\end{equation}
The dynamic scene $\mathcal{S}=(K,\mathbf{g},\Pi,\mathcal{O})$ contains a shared camera model, gravity, support geometry, and persistent objects. Each object has an anonymous identity, semantic descriptor, conditioned mesh $\mathcal{M}_i$, active interval, and pose trajectory $X_i(t)\in\mathrm{SE}(3)$. The executable world $\mathcal{W}$ adds initial states and physical parameters $\Theta$, while $\mathcal{R}_q$ records a query-specific execution. Only $F_{\mathrm{exec}}$ and $F_{\mathrm{ans}}$ receive $q$, allowing one question-agnostic world to serve all query types. All modules run at inference time, with scene constraints inferred from visual evidence rather than benchmark labels.

\begin{table*}[t]
\centering
\small
\begin{tabular}{L{3.15cm}*{8}{C{1.35cm}}}
\toprule
Method
& \multicolumn{2}{c}{Explanatory}
& \multicolumn{2}{c}{Predictive}
& \multicolumn{2}{c}{Counterfactual}
& \multicolumn{2}{c}{Overall} \\
\cmidrule(lr){2-3}\cmidrule(lr){4-5}\cmidrule(lr){6-7}\cmidrule(lr){8-9}
& per ques. & per opt.
& per ques. & per opt.
& per ques. & per opt.
& per ques. & per opt. \\
\midrule
\rowcolor{black!6}
\multicolumn{9}{l}{\textit{Baselines}} \\
Random & 7.19 & 49.33 & 25.40 & 49.71 & 9.75 & 50.02 & 11.26 & 49.69 \\
Blind Gemini-3-Flash & 27.00 & 64.64 & 24.68 & 48.92 & 10.07 & 50.13 & 19.21 & 56.31 \\
\midrule
\rowcolor{black!6}
\multicolumn{9}{l}{\textit{Foundation VLMs}} \\
Qwen3-VL-235B-A22B & 53.19 & 75.84 & 40.40 & 62.12 & 22.17 & 61.79 & 37.52 & 67.92 \\
GLM-4.6V & 39.33 & 73.75 & 63.64 & 66.74 & 24.31 & 58.70 & 36.68 & 66.02 \\
Gemini-3-Flash & 50.32 & 74.66 & 42.71 & 53.46 & 16.63 & 58.38 & 34.32 & 64.97 \\
Gemini-3.1-Pro & 60.90 & 79.17 & 61.76 & 71.28 & 25.37 & 63.50 & 45.47 & 71.06 \\
GPT-4o & 36.76 & 70.48 & 35.64 & 54.98 & 15.57 & 56.54 & 27.29 & 62.44 \\
GPT-5.5 & \textbf{77.21} & \textbf{89.55} & \textbf{85.71} & \textbf{90.69} & \underline{51.33} & \underline{76.74} & \underline{67.24} & \underline{83.66} \\
\midrule
\rowcolor{black!6}
\multicolumn{9}{l}{\textit{Training-Based Methods}} \\
VideoRFT & 16.66 & 61.98 & 39.25 & 47.40 & 15.35 & 54.98 & 19.74 & 57.28 \\
Video-R1 & 32.79 & 71.14 & 41.56 & 46.25 & 19.62 & 59.10 & 28.43 & 63.08 \\
VideoThinker-R1 & 17.53 & 63.60 & 41.56 & 42.50 & 15.14 & 53.71 & 20.37 & 56.92 \\
Chain-of-Frames & 44.71 & 77.26 & \underline{84.42} & \underline{87.88} & 33.48 & 70.84 & 46.21 & 75.29 \\
\midrule
\rowcolor{black!6}
\multicolumn{9}{l}{\textit{Training-Free Methods}} \\
VideoAgent & 34.42 & 61.55 & 14.29 & 48.56 & 7.57 & 51.61 & 19.39 & 55.63 \\
STAR & 55.58 & 78.62 & 24.24 & 57.14 & 7.57 & 53.47 & 29.46 & 64.75 \\
\textbf{PhysMind} & \underline{76.97} & \underline{87.38} & 66.96 & 82.32 & \textbf{70.58} & \textbf{88.08} & \textbf{72.55} & \textbf{87.22} \\
\bottomrule
\end{tabular}
\caption{CLEVRER validation-subset accuracy (\%). Per-question accuracy requires all options of a question to be correct. Bold and underline denote the best and second-best results in each column, respectively.}
\label{tab:clevrer-main}
\end{table*}

\subsection{Dynamic Scene Reconstruction}

Dynamic scene reconstruction estimates persistent identities, canonical geometry, and time-varying poses in a shared 3D coordinate system. The VLM supplies semantics and selects visually supported constraints; specialized tools recover the geometry.

\paragraph{Object Segmentation and Tracking} PhysMind applies SAM~3~\cite{carion2026sam3} with a broad object prompt to obtain anonymous mask tracks $\tau_i=\{M_{i,t}\}_{t=1}^{T}$. For each track, PhysMind selects the frame in which the object is most clearly visible, favoring masks isolated from other tracks. A VLM assigns color, material, geometry, and appearance; these attributes guide mesh conditioning and language grounding, while anonymous identities anchor geometric processing. Visible and fully observed portions define the active and pose-estimation intervals.

\paragraph{Geometry-Aware Object Reconstruction} MoGe-2~\cite{wang2025moge2} estimates intrinsics on sampled frames, which are aggregated into a fixed matrix $K$, and Video Depth Anything~\cite{chen2025videodepth} supplies a temporally consistent depth sequence $\{D_t\}$. At representative frame $k_i$, a masked pixel $\mathbf{u}=(u,v,1)^\top$ is back-projected as
\begin{equation}
\mathbf{P}_{k_i}(\mathbf{u})
=D_{k_i}(\mathbf{u})K^{-1}\mathbf{u}.
\label{eq:backproject}
\end{equation}
The point map, RGB image, and mask condition SAM~3D Objects~\cite{chen2026sam3d} to recover a colored canonical mesh. The VLM-inferred type selects conditioning: regular objects are refitted as primitives, while irregular meshes are simplified with visible surfaces preserved. Visibility-aware rendering refines scale and translation against the mask without changing front-surface depth. Recentering the local frame at the mesh bounds yields a compact proxy for tracking and contact.

\paragraph{Spatiotemporal Pose Recovery and Alignment} FoundationPose~\cite{wen2024foundationpose} registers each conditioned mesh from RGB, depth, mask, and intrinsics. Its forward and backward tracks are joined into a preliminary trajectory $\widetilde{X}_{i,t}=[\widetilde{R}_{i,t}\mid\widetilde{\mathbf{p}}_{i,t}]$, preserving a common object frame while reducing initialization sensitivity.

Because independent tracks may disagree in orientation and support, a question-agnostic VLM determines whether correction is needed and selects visually supported constraints, including shared-surface motion, stable camera roll, and support geometry. The default path robustly aggregates GeoCalib~\cite{veicht2024geocalib} up-direction estimates. When shared-surface motion is detected, the agent invokes support-constrained alignment using visible geometry and dominant motion directions, adding a roll constraint only when supported. The correction respects object symmetries: sphere rotation is ignored, cube and cylinder axes are aligned with gravity when applicable, and rotation about a cylinder's symmetry axis is normalized. Positions are then adjusted along camera rays to restore contact without changing visible surface depth. This joint correction yields trajectories $\{X_i(t)\}$ and support geometry $\Pi$.

\begin{table*}[t]
\centering
\small
\begin{tabular}{L{3.25cm}*{6}{C{1.85cm}}}
\toprule
Method & Fric. Plat. & Fric. Coll. & Bounce Plat. & Bounce Wall & Mass Coll. & Overall \\
\midrule
\rowcolor{black!6}
\multicolumn{7}{l}{\textit{Baselines}} \\
Random & 48.44 & 45.31 & 53.12 & 51.04 & 43.75 & 48.96 \\
Blind Gemini-3-Flash & 46.88 & 50.00 & 47.92 & 52.08 & 51.56 & 49.74 \\
\midrule
\rowcolor{black!6}
\multicolumn{7}{l}{\textit{Foundation VLMs}} \\
Qwen3-VL-235B-A22B & 48.44 & 43.75 & \underline{56.25} & 50.00 & 42.19 & 48.96 \\
Gemini-3-Flash & \textbf{56.25} & 46.88 & 52.08 & 50.00 & 53.13 & 51.56 \\
GPT-5.5 & 51.56 & \textbf{65.62} & 55.21 & \underline{58.33} & \underline{60.94} & \underline{58.07} \\
\midrule
\rowcolor{black!6}
\multicolumn{7}{l}{\textit{Training-Based Methods}} \\
Video-R1 & \underline{54.69} & 53.13 & 51.04 & 45.83 & 53.13 & 51.04 \\
Chain-of-Frames & 48.44 & 51.56 & 50.00 & 51.04 & 50.00 & 50.26 \\
\midrule
\rowcolor{black!6}
\multicolumn{7}{l}{\textit{Training-Free Methods}} \\
\textbf{PhysMind} & \underline{54.69} & \underline{64.06} & \textbf{57.29} & \textbf{59.38} & \textbf{64.06} & \textbf{59.64} \\
\bottomrule
\end{tabular}
\caption{Physion++ object contact prediction per-scene accuracy (\%). Fric., Plat., and Coll. abbreviate friction, platform, and collision. Bold and underline denote the best and second-best results in each column, respectively.}
\label{tab:physionpp-main}
\end{table*}

\subsection{Executable Physical World Modeling}

PhysMind converts $\mathcal{S}$ into an executable 3D world by fitting continuous-time analytic dynamics to the recovered trajectories and meshes. The resulting $\mathcal{W}$ supports inspection, continuation, and intervention.

\paragraph{Analytic Continuous-Time System Identification} Let $\mathbf{z}(t)$ collect the dynamic state of every active object, and let $\Theta$ contain initial velocities, ground friction, pairwise mass ratios, and restitution. PhysMind represents motion by analytic segments separated by instantaneous collision updates:
\begin{equation}
\begin{array}{ll}
\mathbf{z}(t)=\Phi_k(t-t_k;\mathbf{z}_k^{+},\Theta),
&t\in[t_k,t_{k+1}),\\
\mathbf{z}_{k+1}^{+}
=\Delta_{k+1}(\mathbf{z}_{k+1}^{-};\Theta).
\end{array}
\label{eq:analytic-dynamics}
\end{equation}
Here, $\Phi_k$ is the closed-form flow for the active motion mode, and $\Delta_{k+1}$ applies the collision response at $t_{k+1}$. Mesh proximity initializes candidate collision intervals, whose timing and response are refined from trajectories without external collision annotations. For each candidate parameter set, PhysMind evaluates the trajectory in closed form under the corresponding motion mode. Free flight follows the ballistic solution, while supported motion uses exact Coulomb friction with explicit stopping-time handling. A normal collision between objects $i$ and $j$ applies an impulse with scalar magnitude $j_{ij}$ and vector $\mathbf{J}_{ij}$:
\begin{equation}
\begin{array}{rcl}
j_{ij}
&=&
-\frac{(1+e_{ij})(\mathbf{v}_i^{-}-\mathbf{v}_j^{-})^\top\mathbf{n}}
{m_i^{-1}+m_j^{-1}},\\
\mathbf{J}_{ij}
&=&
j_{ij}\mathbf{n}.
\end{array}
\label{eq:analytic-collision}
\end{equation}
where $\mathbf{n}$ is the contact normal and $e_{ij}$ is restitution. The updated velocities initialize the next segment, combining smooth motion and collision discontinuities in a single continuous-time model. Mesh geometry and recovered orientation determine candidate contacts.

Contact-connected objects are optimized jointly with one reference mass per component because collisions identify only mass ratios. Cross-component scales are unobservable without prior contact; new contacts use semantic priors and parameters from visually similar objects. Given corrected positions $\mathbf{p}_{i,t}$ and orientations $R_{i,t}$, identification solves
\begin{equation}
\begin{array}{rcl}
\Theta^{*}&=&\displaystyle\arg\min_{\Theta}
\Bigg[
\sum_{i,t}w_{i,t}
\|\widehat{\mathbf{p}}_i(t;\Theta)-\mathbf{p}_{i,t}\|_2^2\\
&&\displaystyle+\lambda_R\sum_{i,t}w_{i,t}
d_R\!\left(\widehat{R}_i(t;\Theta),R_{i,t}\right)^2\\
&&\displaystyle+\lambda_{\mathrm{contact}}\mathcal{L}_{\mathrm{contact}}(\Theta)
+\lambda_{\mathrm{reg}}\Omega(\Theta)\Bigg].
\end{array}
\label{eq:identification}
\end{equation}
Here, $w_{i,t}$ is observation confidence, $d_R$ is a symmetry-aware rotation distance that discounts ambiguous axes, $\mathcal{L}_{\mathrm{contact}}$ penalizes residual separation and non-approaching motion at candidate contacts, and $\Omega$ is a soft-contact prior; constrained parameterizations enforce physical bounds. Before joint refinement, local analytic fits initialize velocity and friction from observed displacement and deceleration. Insufficiently observed starts use conservative zero-velocity initialization. Coupled objects share one optimization problem, allowing both pre- and post-contact motion to constrain mass ratios and collision response. For long trajectories with frequent direction-changing collisions, optimization uses cumulative temporal prefixes. Each stage extends the horizon from the preceding solution and introduces later collisions progressively before the final full-trajectory fit. Avoiding fixed-step unrolling reduces computation and gradient-path length while preserving collision discontinuities. Appendix~A specifies supported modes, parameterization, bounds, schedules, optimizers, and tool interfaces.

\paragraph{Query-Conditioned Physical Rollout} The question enters only after $\mathcal{W}$ is constructed. A VLM grounds linguistic references to anonymous object identities, determines the required evidence, and composes operations through a general world interface. Explanatory queries inspect states and contacts; predictive queries continue the fitted dynamics; and counterfactual queries clone, edit, and compare worlds. Unless the query requires a different horizon, execution extends 20\% beyond the observed sequence. The unchanged base world can serve multiple questions without repeated reconstruction. Each execution yields $\mathcal{R}_q$, a structured record of trajectories, contacts, state changes, and interventions. The VLM selects the operations and horizon, changing fitted parameters only when the question specifies a physical intervention.

\paragraph{Artifact-Grounded Answering} The answering VLM receives the video and question with compact summaries of $\mathcal{W}$ and $\mathcal{R}_q$, then maps their outcomes to the answer space. Semantic decisions remain with the VLM, while trajectories, fitted dynamics, and executions provide inspectable physical evidence.

\begin{figure*}[t]
\centering
\includegraphics[width=\textwidth]{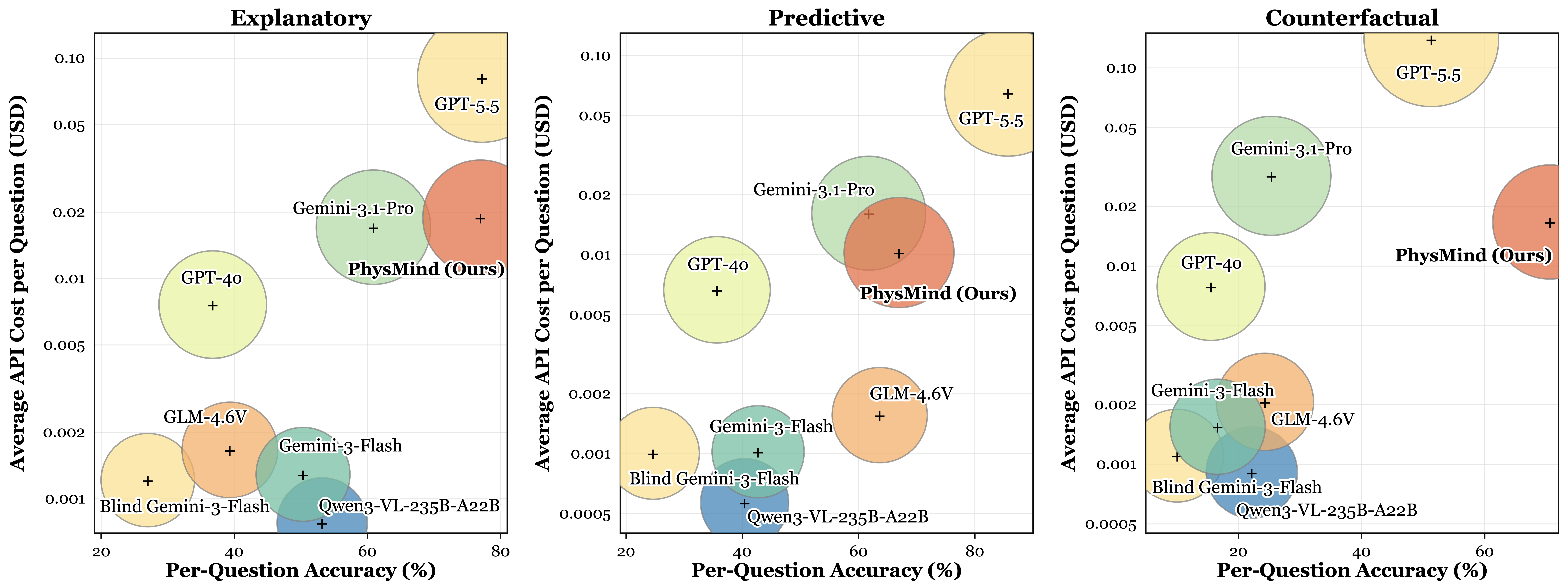}
\caption{API token-cost comparison on CLEVRER. Marker position shows per-question accuracy and average token-derived API cost for each reasoning category; bubble area reflects relative token cost. The vertical axis is logarithmic.}
\label{fig:accuracy-cost}
\end{figure*}

\begin{figure*}[t]
\centering
\includegraphics[width=\textwidth]{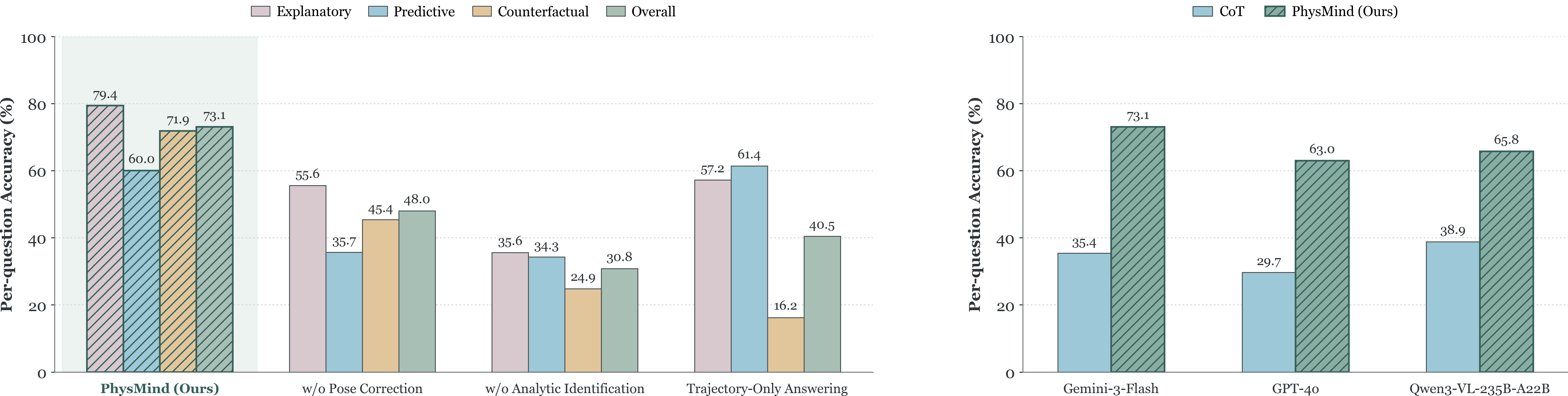}
\caption{Ablation studies on 100 CLEVRER scenes. \textit{(Left)} Component ablations report per-question accuracy for each reasoning category and overall. \textit{(Right)} Agent-VLM ablations compare direct CoT with PhysMind using overall per-question accuracy.}
\label{fig:ablations}
\end{figure*}

\section{Experiments}

This section evaluates PhysMind on CLEVRER~\cite{yi2020clevrer} and Physion++~\cite{tung2023physionpp} against foundation VLMs and representative physical reasoning methods.

\subsection{Experimental Settings}

\paragraph{Benchmarks and Metrics} CLEVRER~\cite{yi2020clevrer} contains synthetic moving-object videos with explanatory, predictive, and counterfactual questions. Under the available compute budget, we use a fixed validation subset of 1,000 scenes, 4,280 questions, and 14,228 options. Per-option accuracy scores binary decisions independently; per-question accuracy requires every option to be correct. Physion++~\cite{tung2023physionpp} evaluates latent-property inference through future contact prediction. We select five rigid-body categories totaling 384 trials and report per-scene accuracy. Every input ends at the benchmark prediction boundary, before outcome frames.

\paragraph{Baselines} We compare four groups: random and blind references; foundation VLMs, including Qwen3-VL-235B-A22B~\cite{bai2025qwen3vl}, GLM-4.6V~\cite{zai2025glm46v}, Gemini-3-Flash~\cite{google2025gemini3flash}, Gemini-3.1-Pro~\cite{google2026gemini31pro}, GPT-4o~\cite{openai2024gpt4o}, and GPT-5.5~\cite{openai2026gpt55}; training-based VideoRFT~\cite{wang2025videorft}, Video-R1~\cite{feng2025videor1}, VideoThinker-R1~\cite{wu2026videothinker}, and Chain-of-Frames~\cite{ghazanfari2025chainofframes}; and training-free VideoAgent~\cite{fan2024videoagent} and STAR~\cite{fan2025star}.

\paragraph{Implementation Details} Gemini-3-Flash is used throughout PhysMind, with one question-agnostic reconstruction shared across a video's questions. No component is trained or fine-tuned on either benchmark, and foundation VLM baselines use CoT. Gemini models receive the observation clip; interfaces without video support receive eight CLEVRER or 16 chronological Physion++ frames. Both video and sampled-frame Physion++ inputs retain the benchmark-provided red/yellow target cues shown after the scene freezes. Appendices~A, B, and~E provide additional implementation and evaluation details.

\begin{figure*}[t]
\centering
\includegraphics[width=0.94\textwidth]{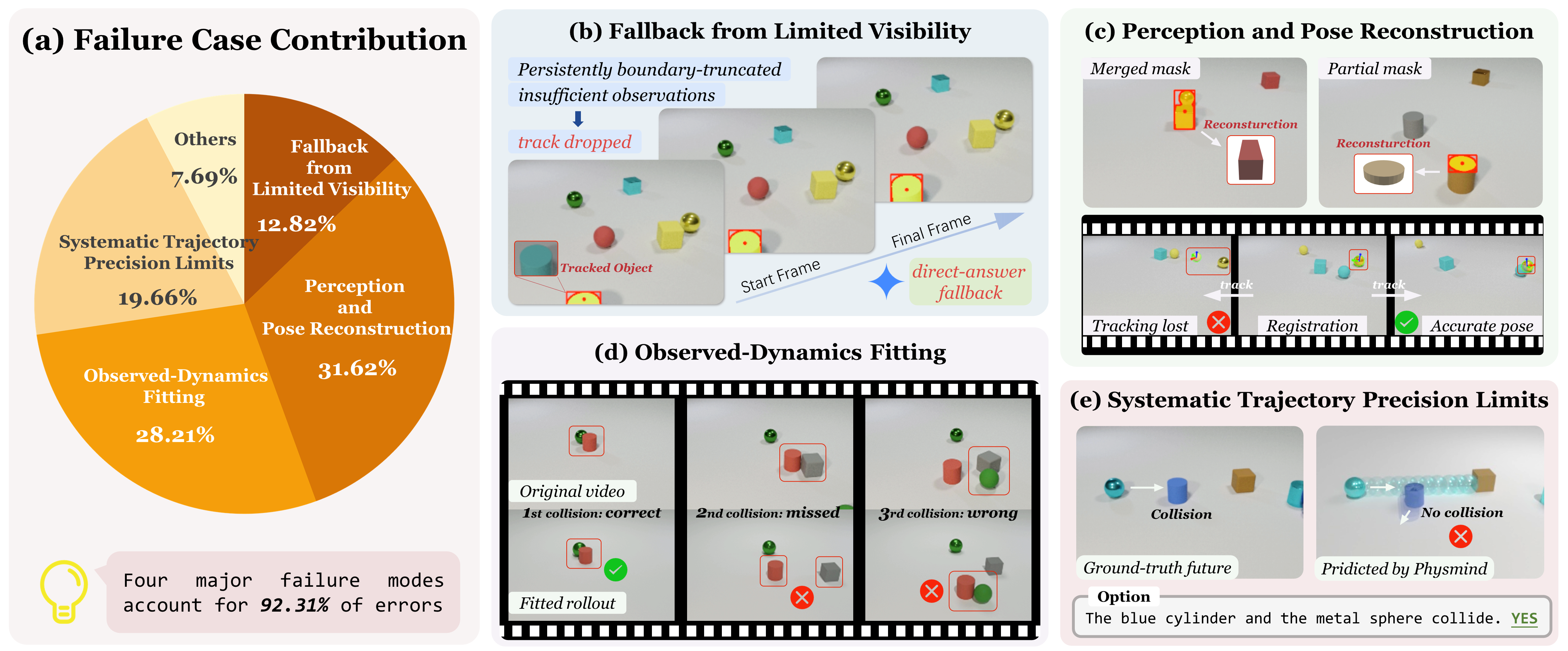}
\caption{Manual error attribution on 100 CLEVRER scenes: (a) category shares and (b--e) representative failure modes.}
\label{fig:error-analysis}
\end{figure*}

\subsection{Main Results}

\paragraph{Results on CLEVRER} Table~\ref{tab:clevrer-main} shows that PhysMind obtains the highest overall accuracy, reaching 72.55\% per question and 87.22\% per option. Relative to same-backbone CoT, the overall gains are 38.23 and 22.25 points; relative to GPT-5.5, they are 5.31 and 3.56 points. The advantage is concentrated in counterfactual reasoning, where PhysMind exceeds same-backbone CoT by 53.95 points per question and GPT-5.5 by 19.25 points. By comparison, PhysMind nearly matches GPT-5.5 on explanation, with a 0.24-point gap, but trails it by 18.75 points on prediction.

This predictive-counterfactual contrast follows the evidence available to each task. Prediction can often continue motion already visible in the video, whereas counterfactual questions require interactions absent from the input, such as collisions after an object is removed. PhysMind evaluates these alternatives by changing the reconstructed state and executing the fitted dynamics. Its larger counterfactual gain is therefore consistent with the executable world providing evidence that cannot be read directly from the observed trajectory.

\paragraph{Results on Physion++} Table~\ref{tab:physionpp-main} reports 59.64\% overall accuracy for PhysMind, improving same-backbone CoT by 8.08 points and GPT-5.5 by 1.57 points. Relative to same-backbone CoT, the gains concentrate on friction collision (17.18 points) and mass collision (10.93 points), while friction-platform accuracy decreases by 1.56 points. This pattern is consistent with executable collision modeling contributing most when the answer depends on pairwise interactions, whereas observed deceleration already supplies direct evidence for platform friction. PhysMind also exceeds GPT-5.5 in four of five categories, with gains of 1.05--3.13 points, and trails it by 1.56 points on friction collision.

\paragraph{Cost--Accuracy Trade-off} Figure~\ref{fig:accuracy-cost} compares CLEVRER per-question accuracy with token-derived API cost. Averaged over the evaluation questions, PhysMind costs \$0.01654 per question. It improves overall accuracy over GPT-5.5 by 5.31 points, while GPT-5.5 costs 6.30 times as much. Compared with Gemini-3.1-Pro, PhysMind improves overall accuracy by 27.08 points and reduces cost by 24.7\%. PhysMind therefore provides the strongest overall cost--accuracy trade-off among these methods.

Appendices~C and~D provide further quantitative results and qualitative cases.

\subsection{Ablation Studies}

Figure~\ref{fig:ablations} shows that the components contribute differently across query types. Replacing analytic identification with fixed-step numerical integration produces the largest overall loss, reducing accuracy from 73.10\% to 30.80\%. Its 42.30-point drop exceeds the losses from removing pose correction (25.05 points) and using trajectories without execution (32.56 points). Analytic identification has a larger effect on explanatory and counterfactual accuracy, which decrease by 43.87 and 47.03 points, than on predictive accuracy, which decreases by 25.71 points. Pose correction produces a more uniform 23.87--26.48-point gain across the three categories, indicating that temporally aligned geometry supports every query type.

Execution is specifically responsible for the counterfactual improvement. Trajectory-only answering slightly increases predictive accuracy from 60.00\% to 61.43\%, but reduces counterfactual accuracy from 71.89\% to 16.22\%. The 55.67-point counterfactual gap separates access to observed motion from the ability to evaluate an intervention. The backbone comparison shows a similar relative pattern: PhysMind improves overall accuracy over direct CoT by 37.7 points with Gemini-3-Flash, 33.3 points with GPT-4o, and 26.9 points with Qwen3-VL-235B-A22B. These consistent within-backbone gains indicate that the executable-world pipeline is not tied to the default agent VLM. Appendix~C reports full results and matched optimization settings.

\subsection{Error Analysis}

We manually assign all 117 errors in the 100-scene subset to the earliest sufficient visible inconsistency. Figure~\ref{fig:error-analysis} distributes them across five categories: perception and pose reconstruction (31.62\%), observed-dynamics fitting (28.21\%), systematic trajectory precision limits (19.66\%), fallback from limited visibility (12.82\%), and others (7.69\%). The four world-construction categories account for 92.31\%. Merged or partial masks co-occur with later pose errors; dynamics errors involve incorrect collision timing or topology; and small trajectory deviations precede long-horizon contact errors. Fallback from limited visibility concentrates near image boundaries and briefly observed objects. Errors after valid artifacts are uncommon. Appendix~D presents representative cases across the evaluated task types.

\section{Conclusion}

PhysMind turns video into a reusable executable physical world without task-specific training. By coordinating dynamic scene reconstruction with analytic continuous-time system identification, it recovers persistent object geometry and motion, fits latent physical properties, and produces query-conditioned rollouts for explanatory, predictive, and counterfactual reasoning. Experiments on CLEVRER and Physion++ show consistent gains over direct CoT and strong VLM baselines, with the largest advantage on counterfactual questions. Component and agent-VLM ablations further support the roles of pose correction, analytic identification, and executable rollout.

\appendix
\section{More Implementation Details}
\label{app:implementation-details}

\subsection{Models and Visual Foundation Modules}
\label{app:models}

PhysMind assembles a modular set of pretrained models to recover a persistent
dynamic scene from video. All modules are used at inference time without
benchmark-specific training or fine-tuning. An agent VLM supplies semantics and
selects scene-dependent constraints and tools, while specialized models provide
geometric outputs for dynamic scene reconstruction and executable physical
world modeling.

\paragraph{Gemini-3-Flash}
We use Gemini-3-Flash~\cite{google2025gemini3flash} as the default agent VLM. It
receives the observation video and supports five semantic functions: compiling
the scene-level object inventory, assigning visual attributes to anonymous
tracks, selecting visually justified reconstruction constraints, planning
query-conditioned rollout operations, and producing the final answer from
execution artifacts. The executable world is constructed once per video and is
shared across its questions, so these question-dependent calls do not repeat
dynamic scene reconstruction.

\paragraph{SAM 3}
SAM~3~\cite{carion2026sam3} performs prompt-based segmentation and tracking over
the complete video. A broad object prompt produces anonymous mask tracks,
after which PhysMind chooses one representative frame for each track by
favoring isolated masks and then larger visible regions. The selected masks
serve multiple downstream stages: they define object-specific reconstruction
inputs, delimit the active and pose-estimation intervals, and initialize
FoundationPose without invoking a second segmentation pass.

\paragraph{MoGe-2}
MoGe-2~\cite{wang2025moge2} estimates camera intrinsics. PhysMind evaluates
eight uniformly sampled frames and takes the componentwise median of their
pixel-space intrinsic estimates to obtain one fixed camera matrix \(K\). This
aggregation reduces frame-specific calibration variation and provides a shared
projection model for depth back-projection, mesh reconstruction, and pose
tracking.

\paragraph{Metric Video Depth Anything}
We run the metric ViT-L variant of Video Depth
Anything~\cite{chen2025videodepth} on the full video. It produces a temporally
consistent metric depth sequence aligned with the fixed camera intrinsics. At
an object's representative frame, the corresponding depth map and mask are
back-projected to form the point map used for object reconstruction.

\paragraph{SAM 3D Objects}
SAM~3D Objects~\cite{chen2026sam3d} reconstructs a colored canonical mesh for
each tracked object. Its inputs combine the representative RGB frame, the
selected SAM~3 mask, the fixed intrinsics, and the metric point map recovered
from video depth. The resulting mesh is subsequently conditioned for physical
modeling: regular shapes can be refitted with compact primitives, whereas
irregular meshes are simplified while preserving their visible geometry. The
conditioned mesh is then expressed in the camera frame and recentered into the
local frame used by pose tracking and contact reasoning.

\paragraph{FoundationPose}
FoundationPose~\cite{wen2024foundationpose} estimates the 6D pose trajectory
from the conditioned mesh, RGB frames, metric depth, fixed intrinsics, and
the reused SAM~3 masks. PhysMind registers the object at a selected
initialization frame and tracks it in both temporal directions over the
object's pose-estimation interval. Joining the forward and backward estimates
retains a common object frame while reducing sensitivity to a single
tracking direction.

\paragraph{GeoCalib}
GeoCalib~\cite{veicht2024geocalib} provides the default camera up-direction
estimates for pose correction, which PhysMind robustly aggregates across frames
to align object orientations and the support plane. When stronger
scene-specific evidence is available, the agent instead invokes
support-constrained alignment using visible geometry and dominant motion
directions, adding a roll constraint only when supported.
GeoCalib therefore remains the default calibration path rather than an
unconditional scene assumption.

\paragraph{Grounding DINO}
We use the base variant of Grounding DINO~\cite{liu2024grounding} as an optional
open-set recognition frontend when direct SAM~3 text prompting is insufficient.
It grounds scenario-specific noun phrases into candidate object boxes, after
which PhysMind filters the detections and converts selected boxes into box or
point prompts for SAM~3 segmentation and tracking. Standard scenes continue to
use SAM~3 text prompts directly, so Grounding DINO supplements rather than
replaces the primary segmentation module.

\subsection{Analytic Continuous-Time System Identification versus Fixed-Step Numerical Integration}
\label{app:analytic-system-identification}

Analytic continuous-time identification and fixed-step integration place
approximation at different stages. Fixed-step models approximate state evolution
on temporal grids, whereas PhysMind evaluates each motion mode in closed form
and treats contacts as state-transition events. These choices affect contact
timing, stopping, optimization, and counterfactual replay under the same physical
parameters. Table~\ref{tab:analytic-vs-discrete} summarizes the comparison.

\begin{table*}[t]
\centering
\small
\setlength{\tabcolsep}{5pt}
\begin{tabular}{L{0.15\textwidth}L{0.395\textwidth}L{0.395\textwidth}}
\toprule
Aspect & Analytic continuous-time system identification & Fixed-step numerical integration \\
\midrule
State propagation
& Direct closed-form evaluation within each active motion mode
& Repeated numerical transitions on a grid with step size $h$ \\
Stopping
& Explicit stopping time under Coulomb friction
& Stopping resolved through successive grid updates \\
Contact timing
& Root localization followed by an instantaneous impulse
& Contact resolved at or within numerical steps, depending on the integrator \\
Optimization path
& Scales mainly with fitted segments and collision events
& Unrolls every integration step over the fitted horizon \\
Primary numerical error
& Event-root tolerance and floating-point evaluation
& Time-discretization error plus contact-resolution error \\
Modeling scope
& Efficient for the supported piecewise analytic modes
& More flexible for dynamics without tractable closed-form segments \\
\bottomrule
\end{tabular}
\caption{Mechanistic comparison of the two system-identification formulations. The table
separates numerical properties from model mismatch: analytic propagation avoids
fixed-step error within supported modes, but remains sensitive to incorrect
parameters, contact modeling, and event topology.}
\label{tab:analytic-vs-discrete}
\end{table*}

\paragraph{Within-mode propagation}
Let $h$ denote the step size of a discrete integrator. A rollout of duration
$T$ applies a numerical transition $N=\lceil T/h\rceil$ times,
\begin{equation}
\mathbf{z}_{n+1}=\Psi_h(\mathbf{z}_n;\Theta),
\qquad t_n=nh,
\label{eq:discrete-rollout}
\end{equation}
where the local approximation made by $\Psi_h$ is repeatedly propagated into
later states. PhysMind instead evaluates a motion segment directly at each
requested time. Free flight follows the ballistic solution stated in the main
paper. For supported planar motion with Coulomb friction, define

\begin{equation}
\begin{array}{rcl}
\tau_i(t)&=&\min\!\left(t-t_k,
\frac{\|\mathbf{v}_{i,k}^{+}\|}{\mu_i g}\right),\\
\mathbf{p}_i(t)&=&\mathbf{p}_{i,k}
+\widehat{\mathbf{v}}_{i,k}^{+}
\left(\|\mathbf{v}_{i,k}^{+}\|\tau_i
-\frac{1}{2}\mu_i g\tau_i^2\right),\\
\mathbf{v}_i(t)&=&\widehat{\mathbf{v}}_{i,k}^{+}
\max\!\left(\|\mathbf{v}_{i,k}^{+}\|-\mu_i g(t-t_k),0\right).
\end{array}
\label{eq:analytic-planar-flow}
\end{equation}
Here, $\widehat{\mathbf{v}}_{i,k}^{+}$ is the post-event direction and
$\tau_i$ clamps motion at the exact stopping time. Dense sampling changes only
the queried output times, not the underlying trajectory. Fixed-step stopping
instead depends on successive updates, so its timing and residual velocity vary
with the step size and integration rule.

\paragraph{Contact timing and discontinuities}
PhysMind localizes pairwise contact at the earliest approaching zero of the
separation between the current closed-form segments,

\begin{equation}
c_{ij}(t)=0,
\qquad \dot{c}_{ij}(t)<0.
\label{eq:analytic-contact-root}
\end{equation}
Here, $c_{ij}(t)$ denotes pairwise separation; the derivative condition excludes
separating roots. The solver refines event time within its valid interval,
applies the mass- and restitution-dependent normal impulse,
and starts new closed-form segments from the post-contact state. A fixed-step
method without continuous localization detects events through grid states.
Coarse steps can shift collisions or admit penetration; smaller steps reduce
this error by increasing transitions. Analytic localization removes this
trade-off for the supported contact model but remains sensitive to incorrect
event topology.

\paragraph{System identification and optimization}
The analytic formulation maps initial velocity, friction, pairwise mass ratios,
and restitution directly to the states at the observed timestamps and evaluates
them under the identification objective defined in the main paper. Its
computational path grows primarily with the number of motion segments and
collision events. A differentiable fixed-step formulation instead
backpropagates through the full chain of $N$ transitions, so longer sequences
lengthen the gradient path and repeatedly expose it to contact branches.
PhysMind further uses cumulative temporal prefixes: early stages fit short
trajectory prefixes, later stages introduce subsequent contacts, and each stage
initializes from the best preceding state. This schedule constrains early motion
before optimizing long collision chains. Analytic propagation shortens the
within-segment path, but it does not eliminate non-smoothness when the inferred
contact topology changes.

\paragraph{Parameterization and solver schedule}
The CLEVRER backend constrains the ground-friction coefficient to
$10^{-5}<\mu_i<1$, each pairwise mass ratio to $0.05<m_i/m_j<20$, and
restitution to $0<e_{ij}<1$ through bounded transformations. Free-contact
fitting uses Adam by default. Its default cumulative-prefix schedule extends the
horizon in 10-frame increments, allows up to 400 updates per prefix, selects the
best state by next-window RMSE, and stops a stage after 100 updates without
improvement. L-BFGS supports the same prefix curriculum, while bounded nonlinear
least squares is available for stages without the curriculum.

\paragraph{Prediction, intervention, and error accumulation}
After identification, PhysMind reuses the fitted parameters to continue the
base world or replay it after an intervention. Removing an object filters it
from the reconstructed world and reruns analytic contact search among the
remaining objects. Since within-mode states are evaluated directly, extending
the horizon does not accumulate time-discretization error at every intermediate
step. The remaining errors arise from recovered poses, fitted parameters,
contact modeling, and omitted physical effects. A fixed-step simulator can in
principle represent richer rotational motion, sustained multi-contact, and
time-varying forces more directly, but its rollout accuracy and cost remain
coupled to the chosen step size. The analytic backend makes a bounded modeling
trade-off: it uses ballistic free flight and, for supported motion, piecewise
straight planar trajectories with constant Coulomb deceleration. Pairwise normal
impulses connect these segments. This scope yields compact, replayable dynamics.

\FloatBarrier

\subsection{Query-Conditioned Physical Rollout Tools}
\label{app:rollout-tools}

PhysMind exposes seven schema-validated tools that convert a grounded question
into targeted physical evidence. The planner calls them over the base rollout
and edited rollouts using registered object identifiers. Rollout-dependent calls
name \texttt{base} or copy an exact \texttt{rollout\_id} returned earlier. The
executor rejects invalid arguments and unknown rollouts, then returns structured
states, trajectories, contacts, event times, or cross-rollout differences.

\paragraph{\texttt{simulate\_edit}}
This tool creates a query-conditioned rollout from an edit type, an internal
object identifier, and an optional start frame. The active registry supports
only \texttt{remove\_object}. It removes the object, replays the fitted analytic
dynamics, and returns a rollout identifier, edit manifest, trajectories,
collisions, in/out events, fit error, and rollout boundaries. The analytic
backend accepts removal only from the object's first active frame.

\paragraph{\texttt{check\_contact}}
This tool checks whether two objects contact or collide in a selected rollout,
optionally within an inclusive frame range. It searches emitted collision events
before using a conservative post-hoc footprint scan. The result reports status,
matched frames, nearest observed distance, and whether an event occurs beyond
the source video. For edited rollouts, it compares the first match with the base
rollout under a five-frame tolerance. Distance alone does not establish contact.

\paragraph{\texttt{get\_event\_time}}
This tool retrieves contact or collision times for two objects in a selected
rollout and optional inclusive frame range. It returns all matched frames, event
count, first frame, and first event record, using the same conservative scan
when no collision record matches. The first event supports temporal ordering in
explanatory questions.

\paragraph{\texttt{compare\_rollouts}}
This tool compares reference and edited rollouts, defaulting to the base and
available edited worlds when identifiers are omitted. An optional object list
restricts the comparison. It returns collision pairs in each rollout, pairs
added or removed by the edit, and per-object trajectory differences over shared
frames, including sample count and mean and maximum position change.

\paragraph{\texttt{inspect\_world\_state}}
This tool returns active object records and collision events at one required
video frame. If the trajectory lacks that frame, it returns
\texttt{missing\_frame} rather than interpolating a state.

\paragraph{\texttt{inspect\_object\_trajectory}}
This tool extracts one object's corrected or simulated trajectory using a
required object identifier and optional rollout identifier and inclusive frame
range. It returns the matched per-frame records, their count, and the first and
last records.

\paragraph{\texttt{remove\_object}}
This legacy tool provides a flat interface for object removal. It accepts an
object identifier and optional start frame, then uses the same analytic replay
path and returns the same evidence as \texttt{simulate\_edit}. New
counterfactual plans use \texttt{simulate\_edit} as the general entry point.

\section{Evaluation Benchmark Details}
\label{app:benchmark-details}

\subsection{CLEVRER}
\label{app:clevrer-details}

CLEVRER~\cite{yi2020clevrer} consists of synthetic videos rendered from
physics-engine simulations of a flat tabletop. Each scene contains a small set
of visually simple rigid objects distinguished by color, shape, and material.
They can enter the camera view, move across the surface, collide with one
another, and leave the view. The controlled camera and environment reduce
irrelevant visual variation while retaining nontrivial event sequences and
collision chains. Questions refer to objects through their attributes, making
persistent identity and temporal event order prerequisites for reasoning.

We evaluate a fixed validation subset containing 1,000 scenes, 4,280 questions,
and 14,228 answer options across three multi-select question types. Each answer
option is judged independently, so a question may have zero, one, or multiple
correct options. The source benchmark associates each five-second video with
object motion traces and event histories, whereas PhysMind receives only the
rendered video and language questions. Multiple questions can share one video,
probing causes, future outcomes, and edited outcomes of the same observed event
sequence.

\paragraph{Explanatory questions.}
These questions identify which earlier entering, leaving, or collision events
are responsible for a specified observed event. Correct answers therefore
require recovering the event graph and distinguishing genuine causal
preconditions from events that merely occurred earlier. Because candidate
causes are presented as separate options, the model must retain every
supported causal link instead of returning only the most salient event. A
cause may also be indirect: an early collision can redirect an object that
later participates in the queried event, requiring the intervening collision
chain to be preserved.

\paragraph{Predictive questions.}
These questions ask which events will occur after the observed video ends.
They require continuing the inferred dynamics beyond the observation window
and checking candidate entering, leaving, and collision events against the
continued trajectories. The target evidence is consequently absent from the
input and cannot be recovered through event recognition alone. Several
candidate events can be jointly compatible with one future, so prediction
cannot be reduced to selecting a single next event.

\paragraph{Counterfactual questions.}
These questions evaluate an edited scene in which a specified object is
removed. The altered world must be rolled out again because removing one
object can suppress direct contacts and change downstream collision chains. We
must compare events in this edited trajectory, rather than reuse outcomes from
the factual video or simply delete one event from its event history.

\paragraph{Metrics.}
Across all three question types, we report both per-option accuracy and the
stricter per-question accuracy, which counts a question as correct only when
every option is classified correctly.

\subsection{Physion++}
\label{app:physionpp-details}

Physion++~\cite{tung2023physionpp} tests whether a model can infer latent
mechanical properties from observed motion and use them for future contact
prediction. Our evaluation contains 384 trials from five rigid-body categories:
64 each from Mass Collision, Friction Collision, and Friction Platform, and 96
each from Bounce Platform and Bounce Wall. Every trial poses the same binary
object contact prediction question: if the scene continues beyond the supplied
clip, will the red agent contact the yellow patient?

These categories use two temporal formats. Friction Platform and Bounce
Platform contain no curtain: evidence about the latent property and the motion
to be continued occur in one uninterrupted configuration. Mass Collision, Friction
Collision, and Bounce Wall instead separate inference and prediction with a
curtain. Motion before the curtain reveals mass, friction, or bounciness; while
the scene is occluded, objects are rearranged; and the post-curtain motion
defines the new prediction state. The latent property persists across this
transition, but pre-curtain positions and trajectories do not. In both formats,
the red and yellow cues identify the queried pair without changing their
dynamics. Our input stops at the benchmark prediction boundary and never
exposes outcome frames.

\paragraph{Mass Collision.}
A rolling ball first collides with an object, revealing evidence about its
mass. After a curtain-covered rearrangement, another collision may displace the
red agent from the path of the falling yellow patient; predicting contact
depends on transferring the inferred mass to the new configuration.

\paragraph{Friction Collision.}
The initial sliding motion reveals the red agent's friction. Following a
curtain transition, the agent is reset with a new velocity while the yellow
patient falls from above. Contact occurs only if friction slows the agent into
the patient's landing path rather than causing it to under- or overshoot.

\paragraph{Friction Platform.}
This is a continuous, no-curtain scene in which the agent traverses a
ramp-and-platform arrangement. Its observed deceleration provides the friction
evidence needed to determine whether it stops short of, reaches, or passes the
yellow target region.

\paragraph{Bounce Platform.}
Also without a curtain, the red agent bounces and slides along a raised
platform before falling toward a lower surface. Its observed rebound behavior
reveals bounciness, which determines whether the subsequent trajectory reaches
the yellow target across the intervening platform geometry.

\paragraph{Bounce Wall.}
Before the curtain, an airborne object strikes a wall and rebounds, exposing
its bounciness. The rearranged prediction phase launches the same object toward
a wall again; the transferred restitution determines whether its rebound
brings the red agent into contact with the yellow patient.

\section{More Experimental Results}
\label{app:more-experimental-results}

\subsection{Additional Physion++ Results}
\label{app:physionpp-additional-results}

Table~\ref{tab:physionpp-paired-results} reports per-scene and matched-pair
accuracy for Qwen3-VL-235B-A22B~\cite{bai2025qwen3vl},
Gemini-3-Flash~\cite{google2025gemini3flash},
GPT-5.5~\cite{openai2026gpt55}, Video-R1~\cite{feng2025videor1}, and
Chain-of-Frames~\cite{ghazanfari2025chainofframes}, together with random and
blind references. A matched pair contains two trials whose objects begin the
prediction phase from the same placement but whose latent mechanical property
and contact outcome differ; it is correct only when both trials are answered
correctly. PhysMind leads overall with 59.64\% per-scene and 27.09\% per-pair
accuracy. Its per-pair accuracy exceeds GPT-5.5 by 4.69 points and random
guessing by 4.17 points, indicating more consistent discrimination of the two
property-conditioned outcomes.

To test whether the terminal red/yellow cue is too brief for reliable target
identification, we rerun a subset of baselines, excluding PhysMind, with a red
overlay continuously tracking the agent, a yellow overlay continuously tracking
the patient, and a matching prompt
(Table~\ref{tab:physionpp-tracking-cue}). The changes are mixed:
Gemini-3-Flash improves modestly overall, with larger gains confined to some
scenarios (on Friction Platform, $+9.38$ scene and $+28.13$ pair points),
whereas Video-R1, Chain-of-Frames, and Qwen3-VL-235B-A22B show changes of
$0.00$, $-1.56$, and $-1.04$ overall scene points, respectively. Thus poor
baseline performance cannot generally be attributed to an insufficient
terminal cue; continuously marking both targets still provides no consistent
improvement.

\subsection{Additional CLEVRER Ablations}
\label{app:clevrer-additional-ablations}

Table~\ref{tab:clevrer-additional-ablations} evaluates three isolated component
substitutions on the same 100-scene CLEVRER~\cite{yi2020clevrer} subset used for
the main ablation study: GeoCalib replaces the projection-overlap up-direction
search, GoTrack replaces FoundationPose for pose tracking, and the third
variant removes depth-aware occlusion filtering from mask-IoU matching.

GoTrack~\cite{nguyen2025gotrack} is a generic CAD-based 6D pose refiner that
aligns rendered object templates to RGB frames through flow-based
correspondences. In this ablation, FoundationPose still initializes each
object, while GoTrack recursively refines the preceding pose for temporal
tracking. Thus, unlike FoundationPose's RGB-D render-and-compare tracking,
GoTrack performs RGB-only correspondence-based updates without observed depth.

GeoCalib and GoTrack remain within 1.60 overall points of PhysMind, supporting
robustness to alternative up-direction estimation and pose tracking components.
Removing depth-aware filtering improves all eight metrics, including gains of
3.45 per-question and 1.94 per-option points overall. We interpret this as a
CLEVRER-specific effect: with limited severe occlusion, depth or alignment
noise can make filtering discard useful pixels. Under heavier occlusion,
however, an unfiltered rendered silhouette includes hidden regions absent from
the visible mask, causing systematic mismatch during mesh conditioning and pose
correction. Because this failure mode is relevant to many Physion++ scenes,
PhysMind retains depth-aware filtering as the more robust default.

\section{Qualitative Case Studies}
\label{app:qualitative-case-studies}

\begin{figure*}[p]
\centering
\includegraphics[width=\textwidth,height=0.88\textheight,keepaspectratio]{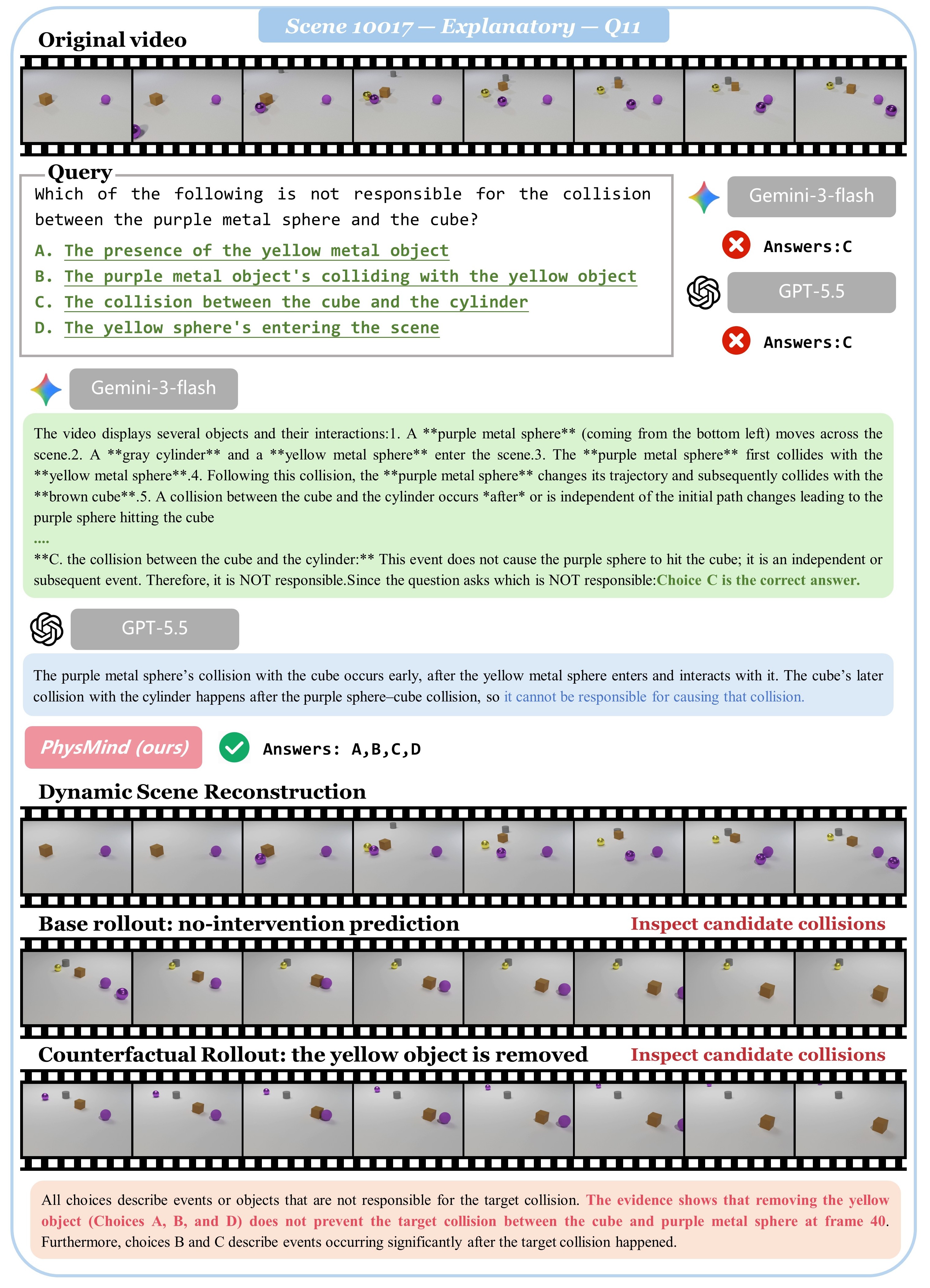}
\caption{CLEVRER explanatory case study for scene 10017, question 11. The
direct VLM baselines select only choice C, whereas PhysMind uses event timing
and intervention evidence to identify all four choices as not responsible for
the target collision.}
\label{fig:case-clevrer-explanatory}
\end{figure*}

\begin{figure*}[p]
\centering
\includegraphics[width=\textwidth,height=0.88\textheight,keepaspectratio]{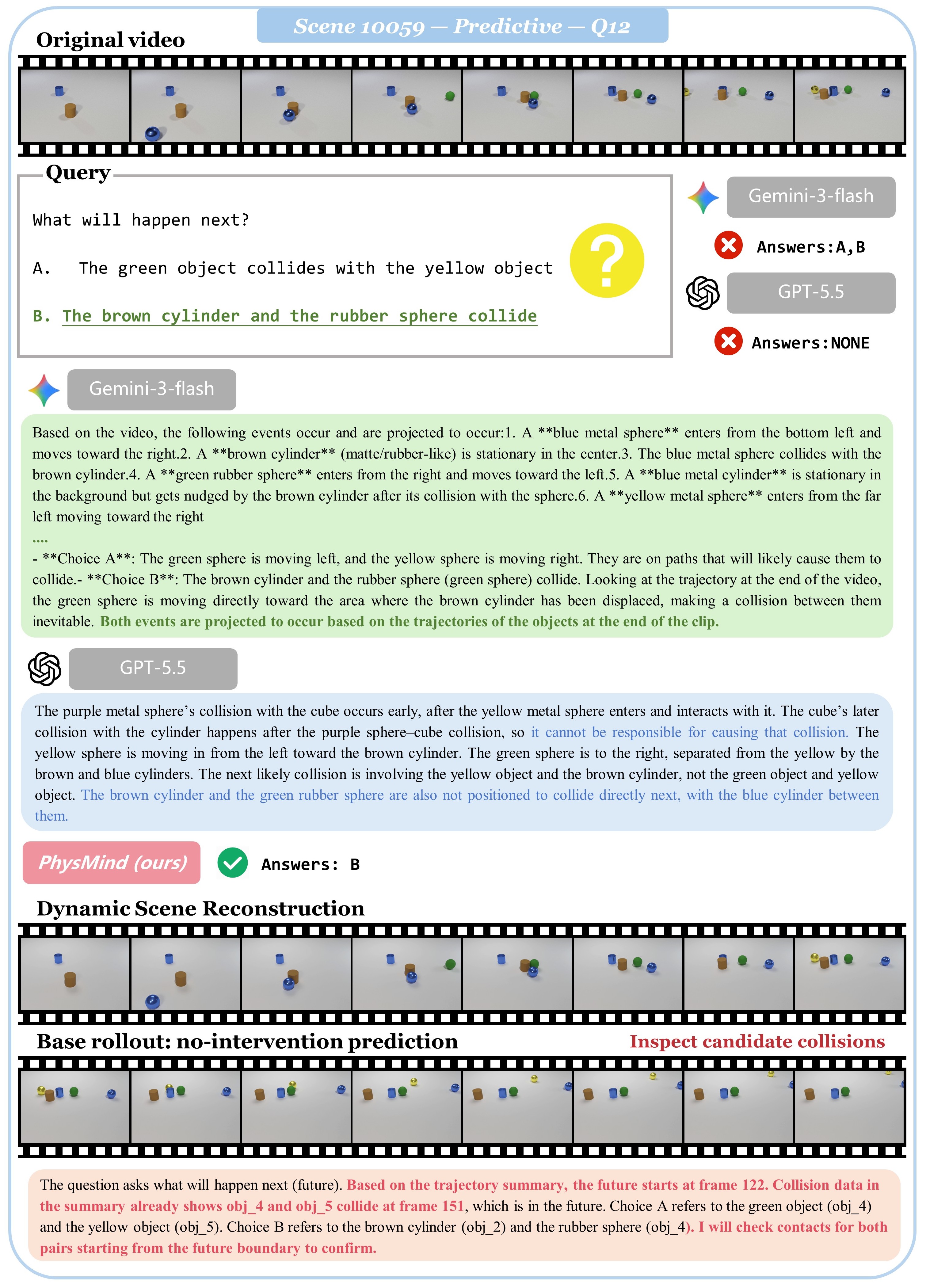}
\caption{CLEVRER predictive case study for scene 10059, question 12. PhysMind
continues the reconstructed world and inspects candidate contacts before
selecting choice B; Gemini-3-Flash selects A and B, while GPT-5.5 selects no
choice.}
\label{fig:case-clevrer-predictive}
\end{figure*}

\begin{figure*}[p]
\centering
\includegraphics[width=\textwidth,height=0.88\textheight,keepaspectratio]{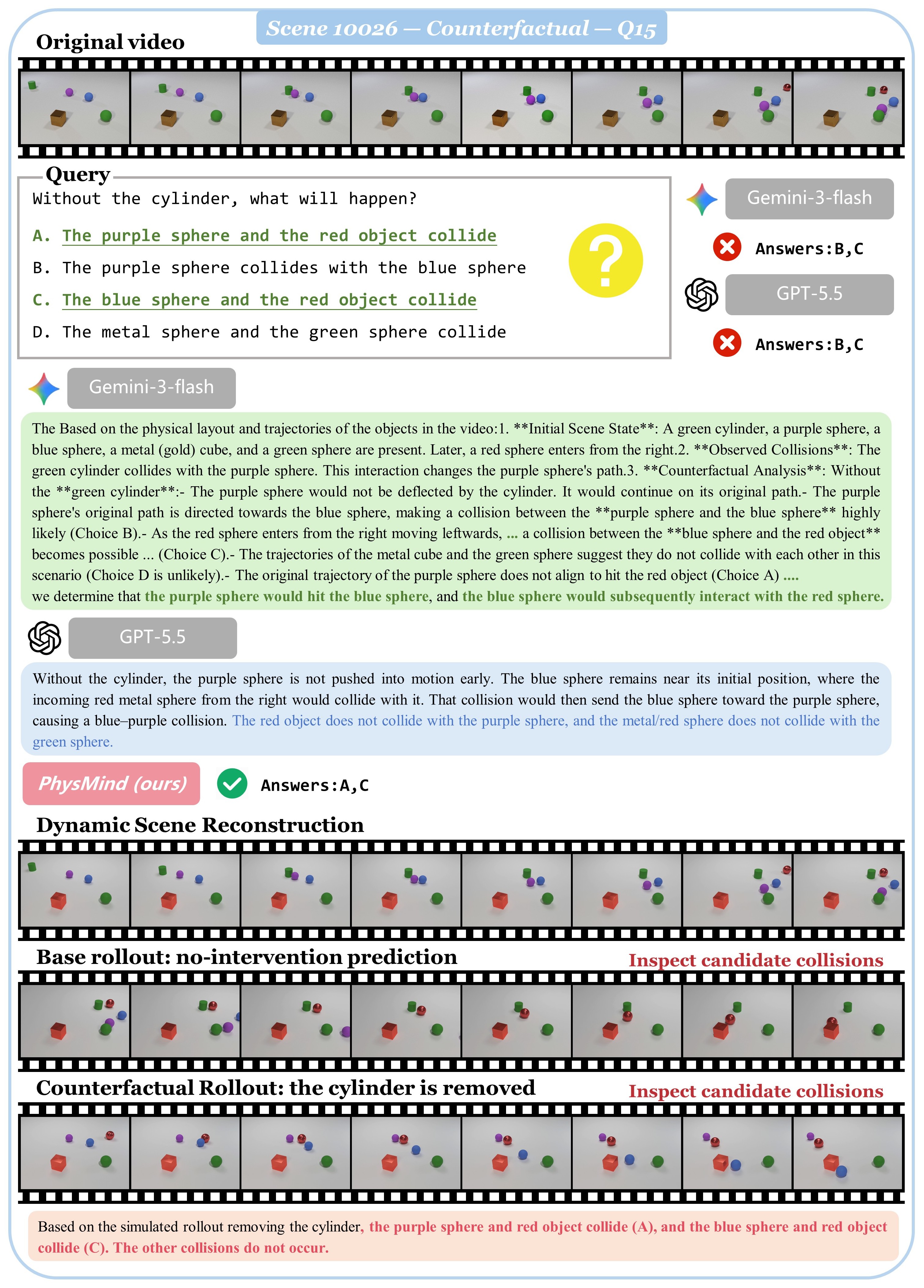}
\caption{CLEVRER counterfactual case study for scene 10026, question 15. After
removing the cylinder, PhysMind's edited rollout supports choices A and C,
whereas both direct VLM baselines select B and C.}
\label{fig:case-clevrer-counterfactual}
\end{figure*}

\begin{figure*}[p]
\centering
\includegraphics[width=\textwidth,height=0.88\textheight,keepaspectratio]{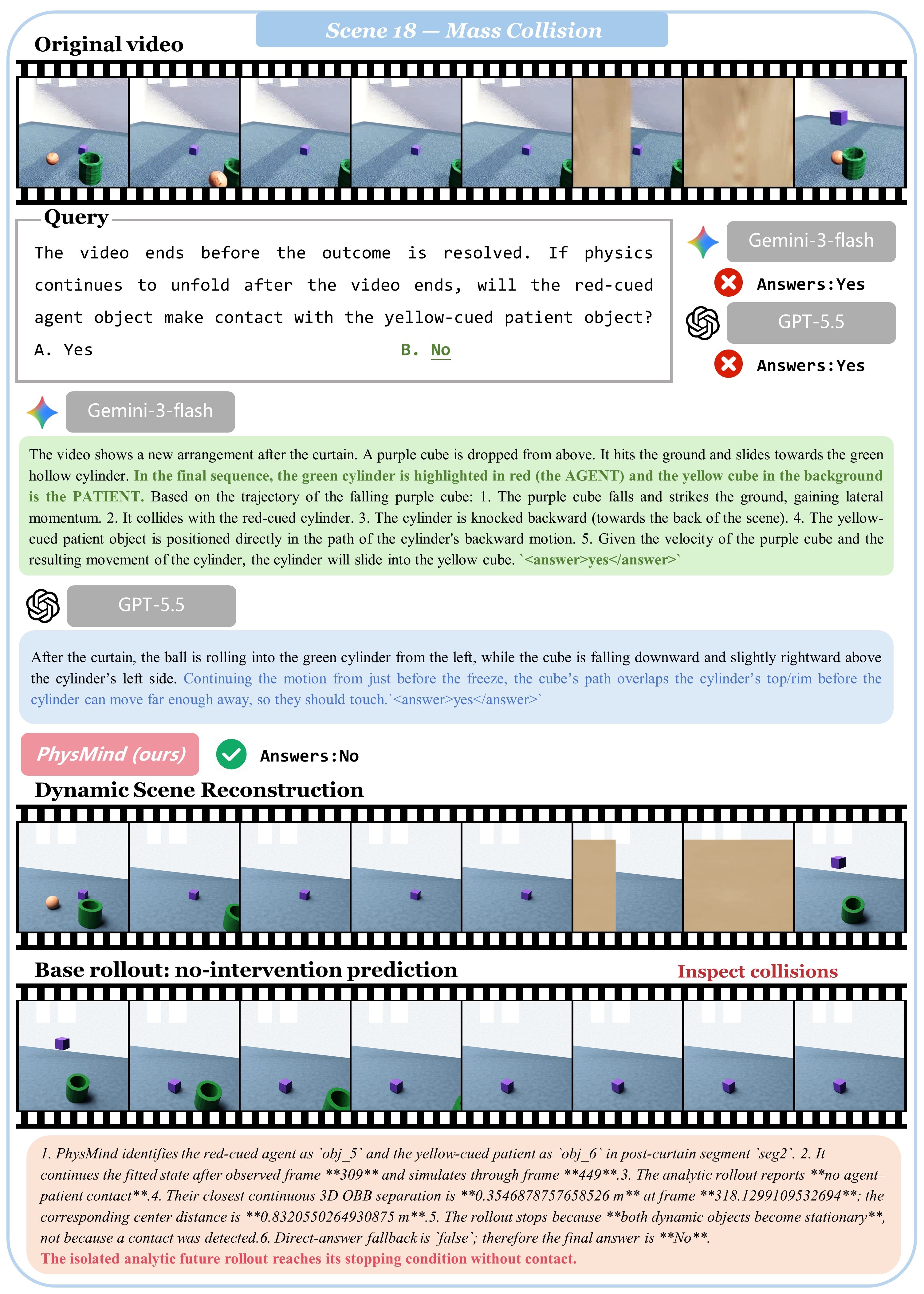}
\caption{Physion++ Mass Collision case study for scene 18. Both direct VLM
baselines predict contact, while PhysMind's analytic rollout reaches its
stopping condition without contact between the red agent and yellow patient.}
\label{fig:case-physion-mass}
\end{figure*}

\begin{figure*}[p]
\centering
\includegraphics[width=\textwidth,height=0.88\textheight,keepaspectratio]{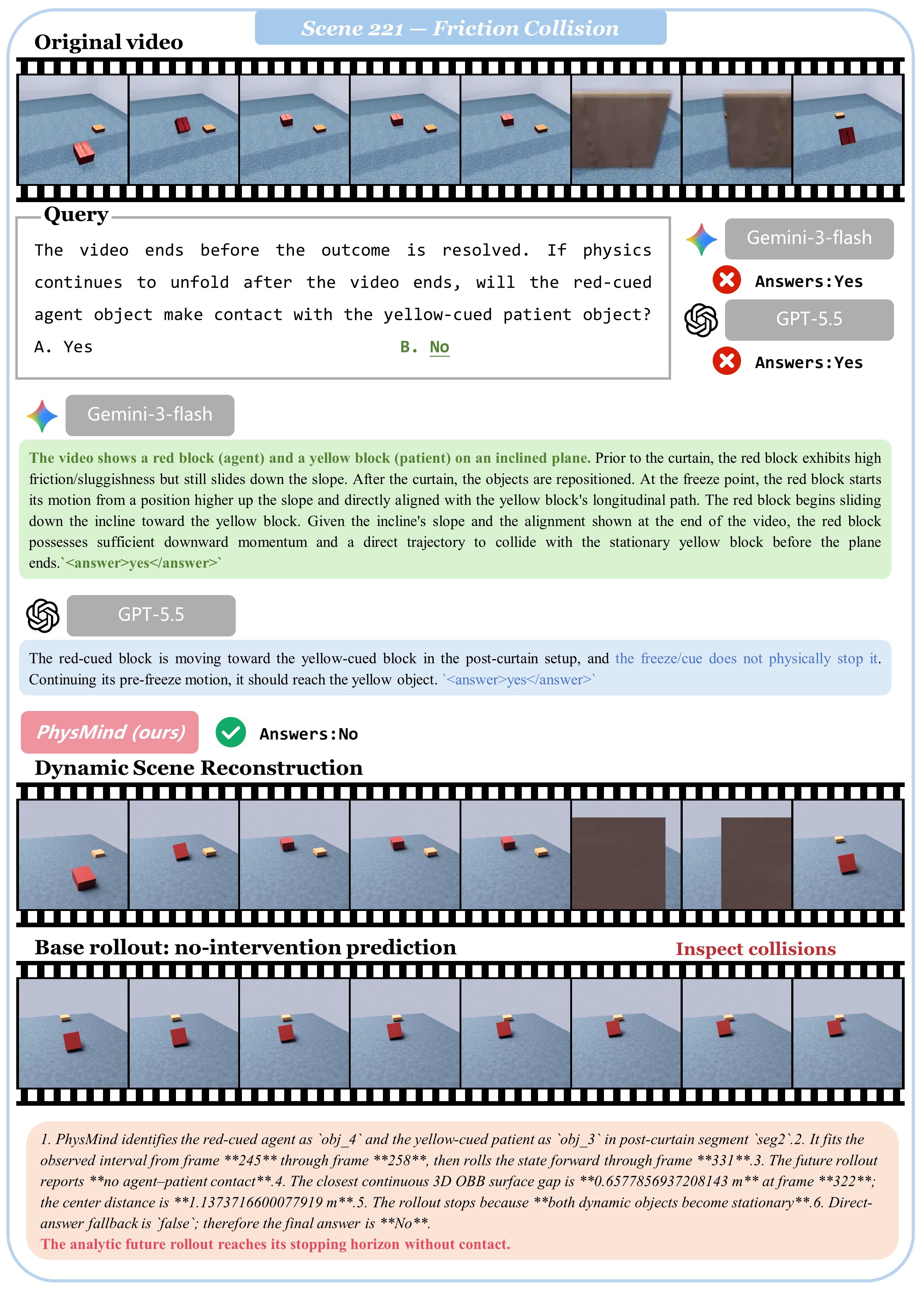}
\caption{Physion++ Friction Collision case study for scene 221. Both direct VLM
baselines predict contact, while PhysMind fits the observed motion and rolls the
post-curtain state forward without contact.}
\label{fig:case-physion-friction-collision}
\end{figure*}

\begin{figure*}[p]
\centering
\includegraphics[width=\textwidth,height=0.88\textheight,keepaspectratio]{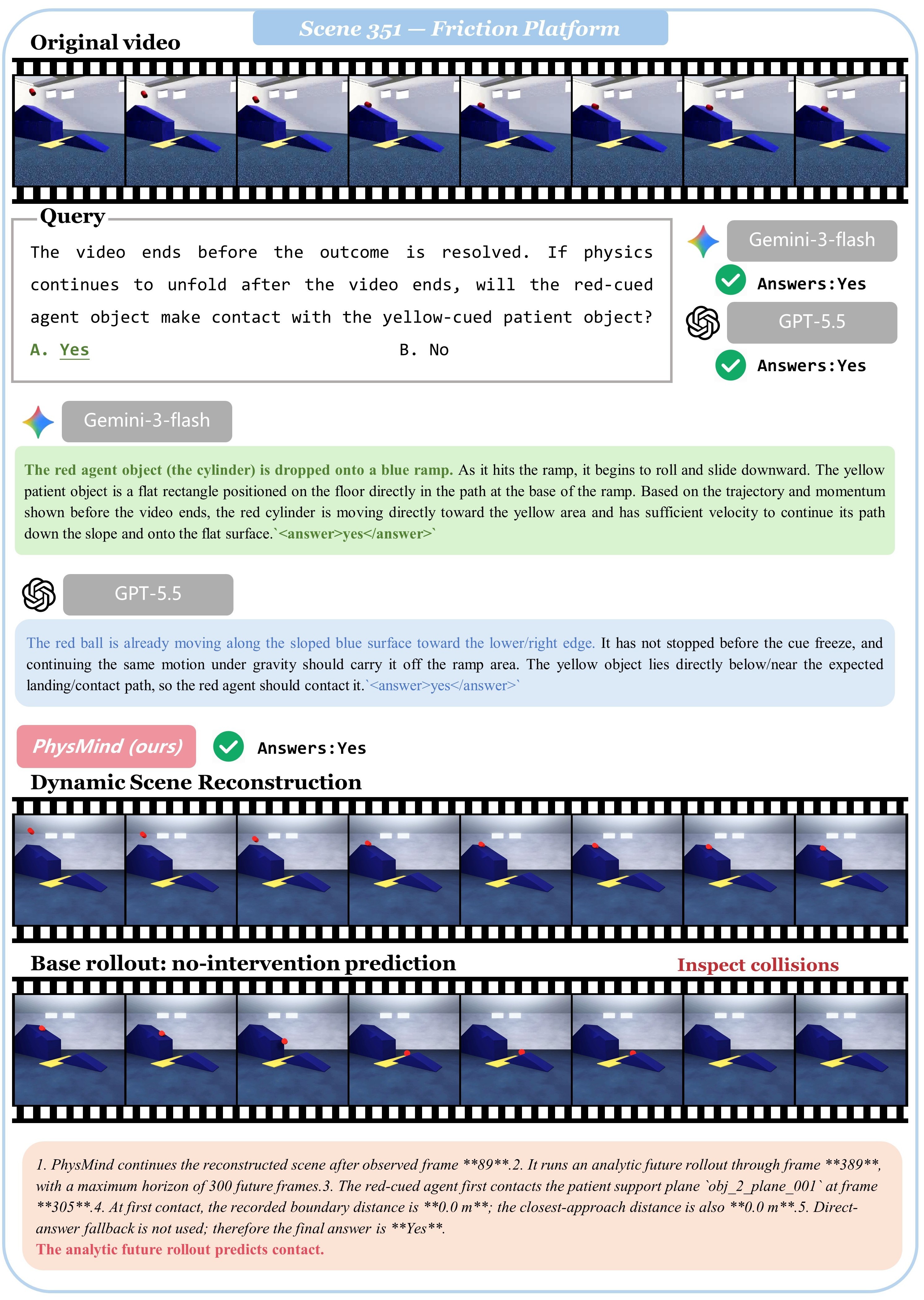}
\caption{Physion++ Friction Platform case study for scene 351. All three
methods predict contact. PhysMind grounds its answer in an analytic continuation
of the reconstructed trajectory.}
\label{fig:case-physion-friction-platform}
\end{figure*}

\begin{figure*}[p]
\centering
\includegraphics[width=\textwidth,height=0.88\textheight,keepaspectratio]{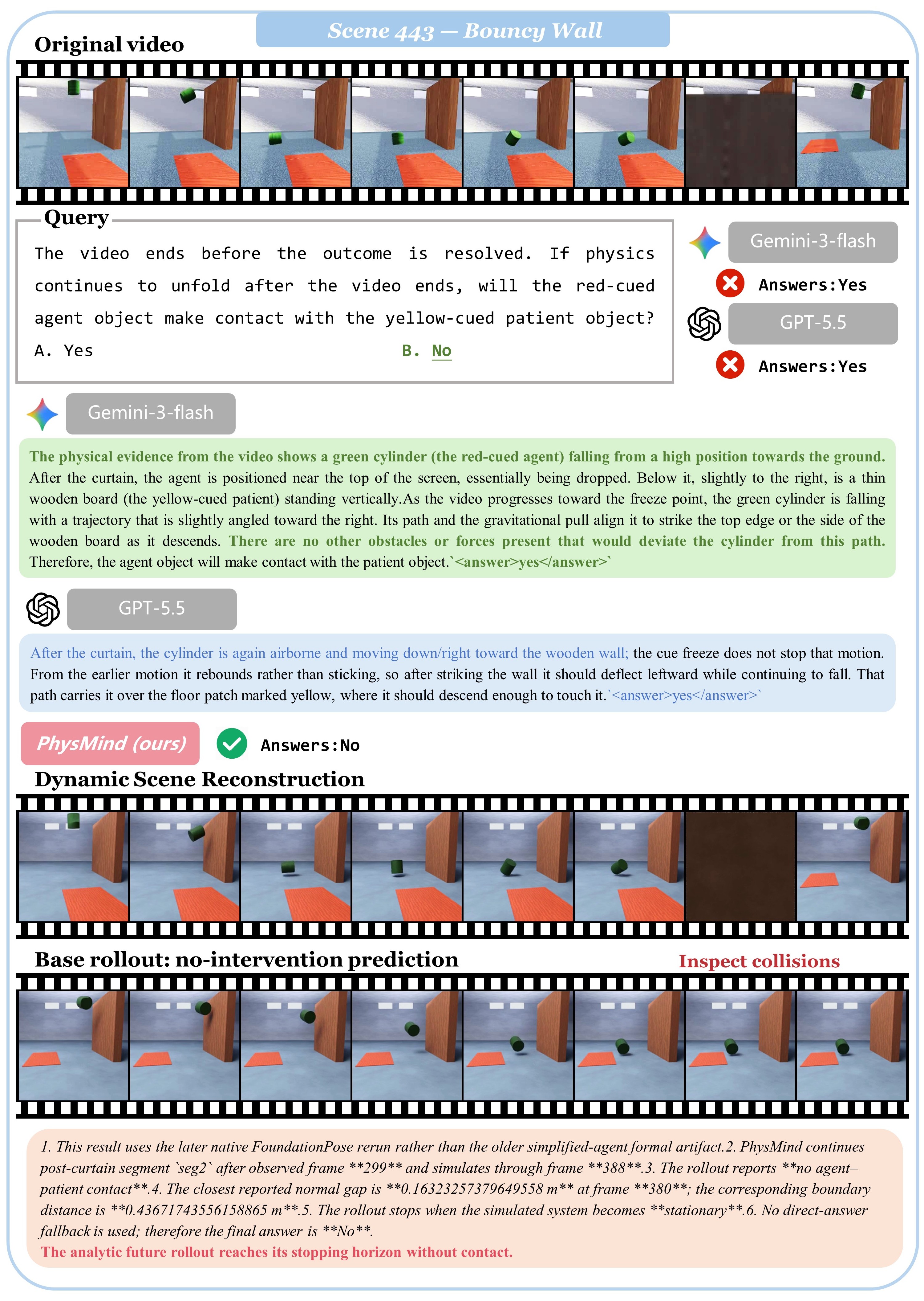}
\caption{Physion++ Bounce Wall case study for scene 443. Both direct VLM
baselines predict contact, while PhysMind's fitted rollout becomes stationary
without contact between the red agent and yellow patient.}
\label{fig:case-physion-bounce-wall}
\end{figure*}

\begin{figure*}[p]
\centering
\includegraphics[width=\textwidth,height=0.88\textheight,keepaspectratio]{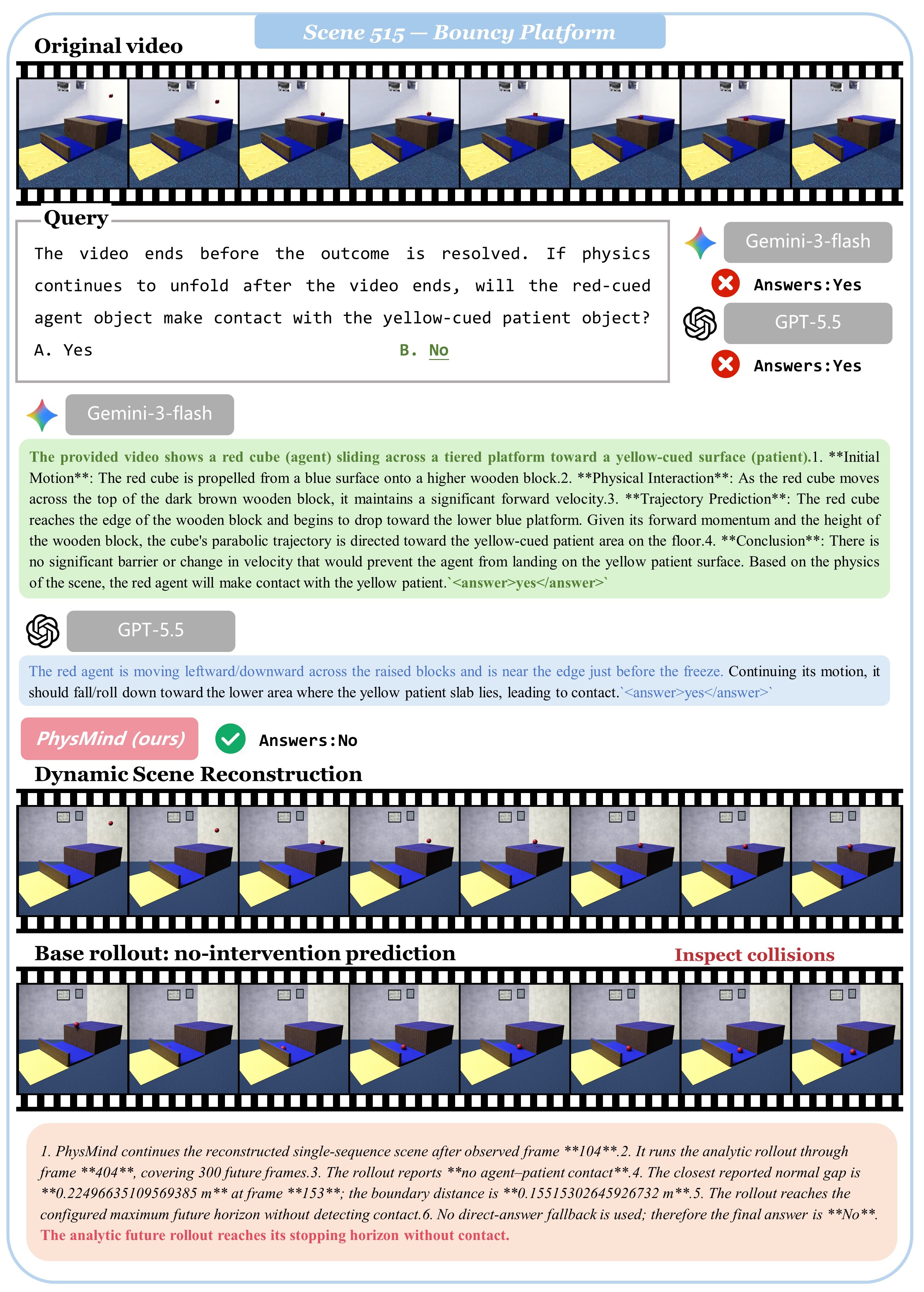}
\caption{Physion++ Bounce Platform case study for scene 515. Both direct VLM
baselines predict contact, while PhysMind's analytic rollout reaches the
configured horizon without contact.}
\label{fig:case-physion-bounce-platform}
\end{figure*}

Figures~\ref{fig:case-clevrer-explanatory}--\ref{fig:case-clevrer-counterfactual}
present explanatory, predictive, and counterfactual CLEVRER cases.
Figures~\ref{fig:case-physion-mass}--\ref{fig:case-physion-bounce-platform}
cover the Mass Collision, Friction Collision, Friction Platform, Bounce Wall,
and Bounce Platform categories in Physion++. Each figure juxtaposes the
observed video, query, and direct VLM reasoning with PhysMind's reconstructed
scene and execution artifacts.

\section{Prompts Used in PhysMind}
\label{app:prompts}

\begin{figure*}[p]
\centering
\includegraphics[width=\textwidth,height=0.84\textheight,keepaspectratio]{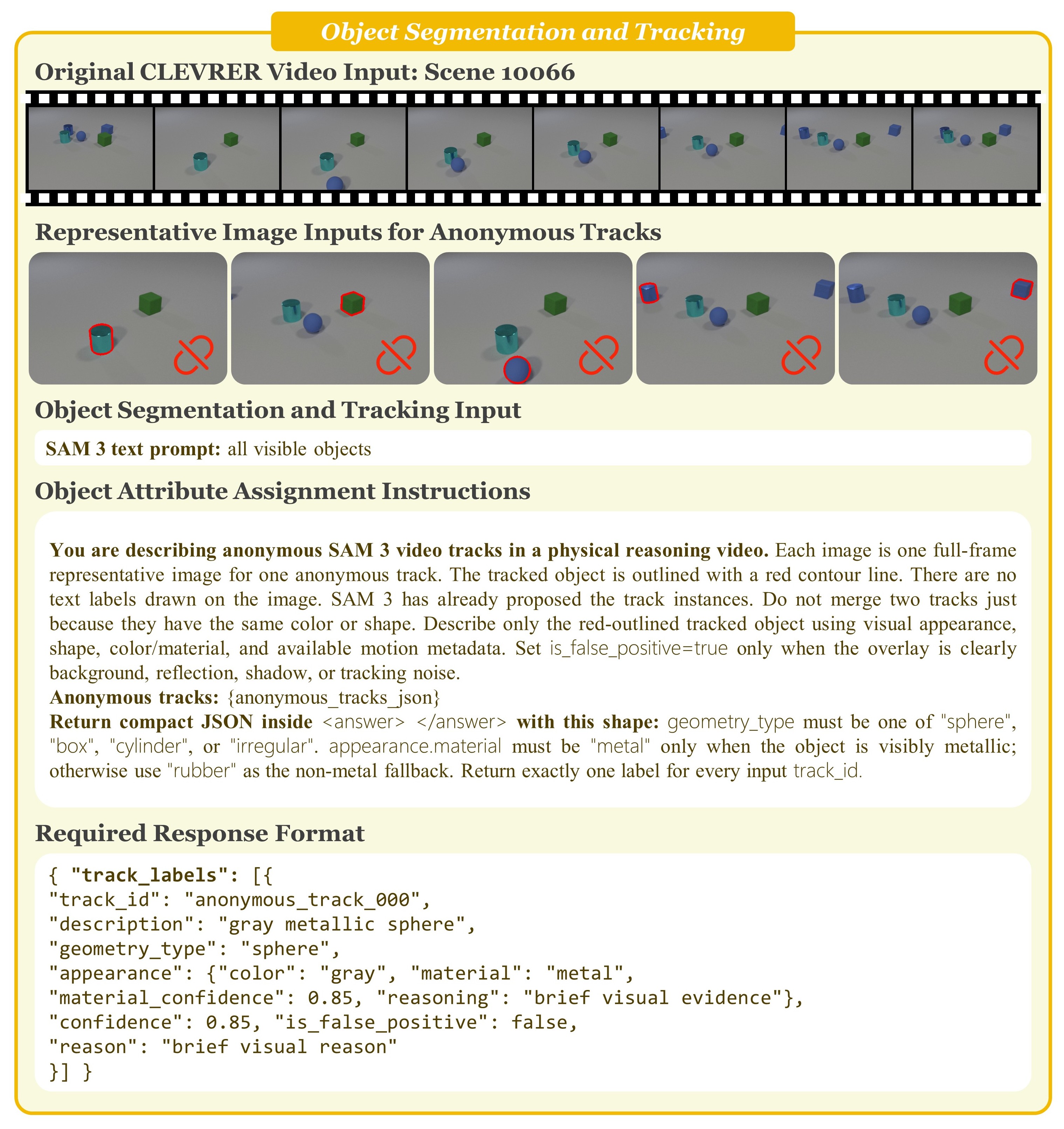}
\caption{Object segmentation and tracking. SAM~3 receives the complete video
and a fixed text prompt. The VLM that assigns color, material, geometry, and
appearance receives one full-frame red-contour representative image per
anonymous track, the injected track metadata, and the required JSON response
format.}
\label{fig:prompt-segmentation}
\end{figure*}

\begin{figure*}[p]
\centering
\includegraphics[width=\textwidth,height=0.84\textheight,keepaspectratio]{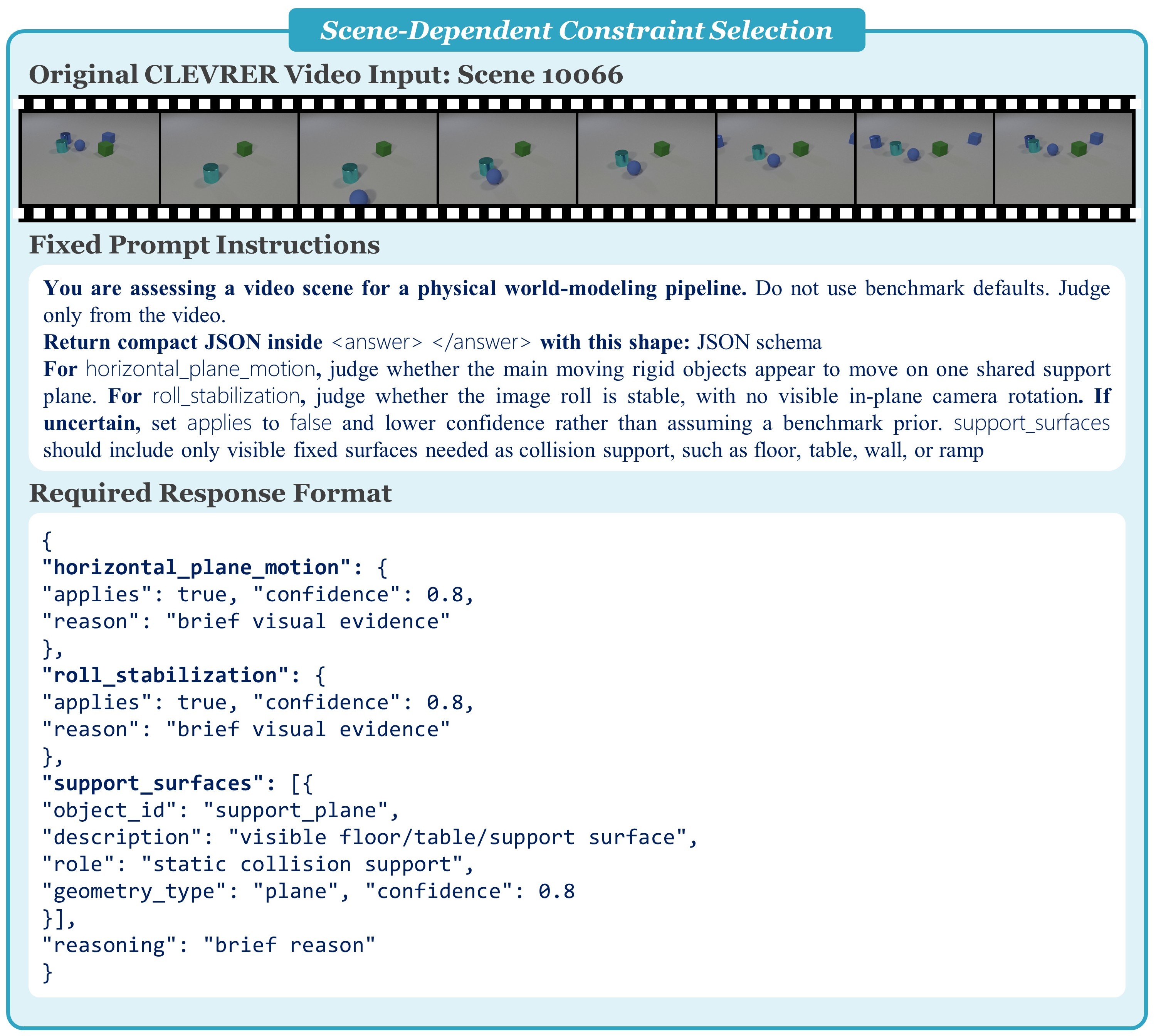}
\caption{Scene-dependent constraint selection. The complete video, fixed
visual-assessment instructions, and required JSON response format form the
model input. The response format constrains the requested assessment; no
generated response is shown.}
\label{fig:prompt-constraint-selection}
\end{figure*}

\begin{figure*}[p]
\centering
\includegraphics[width=\textwidth,height=0.84\textheight,keepaspectratio]{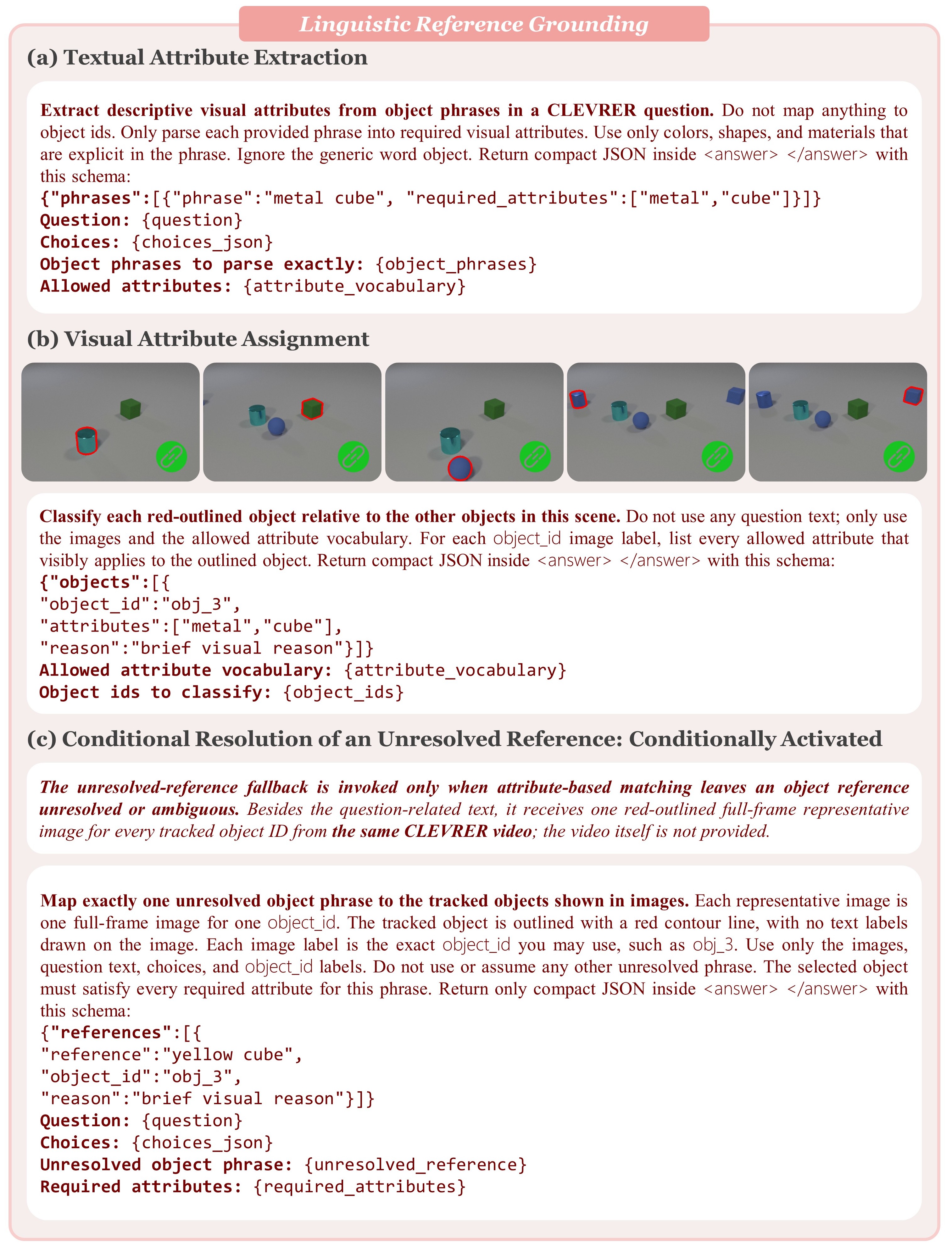}
\caption{Linguistic reference grounding. The first input is text-only and
extracts explicit attributes without assigning identities. The second input
associates visual attributes with red-contour object images. The third input,
used only for unresolved references, supplies the question, choices, required
attributes, and candidate images to request one object identity.}
\label{fig:prompt-reference-grounding}
\end{figure*}

\begin{figure*}[p]
\centering
\includegraphics[width=\textwidth,height=0.84\textheight,keepaspectratio]{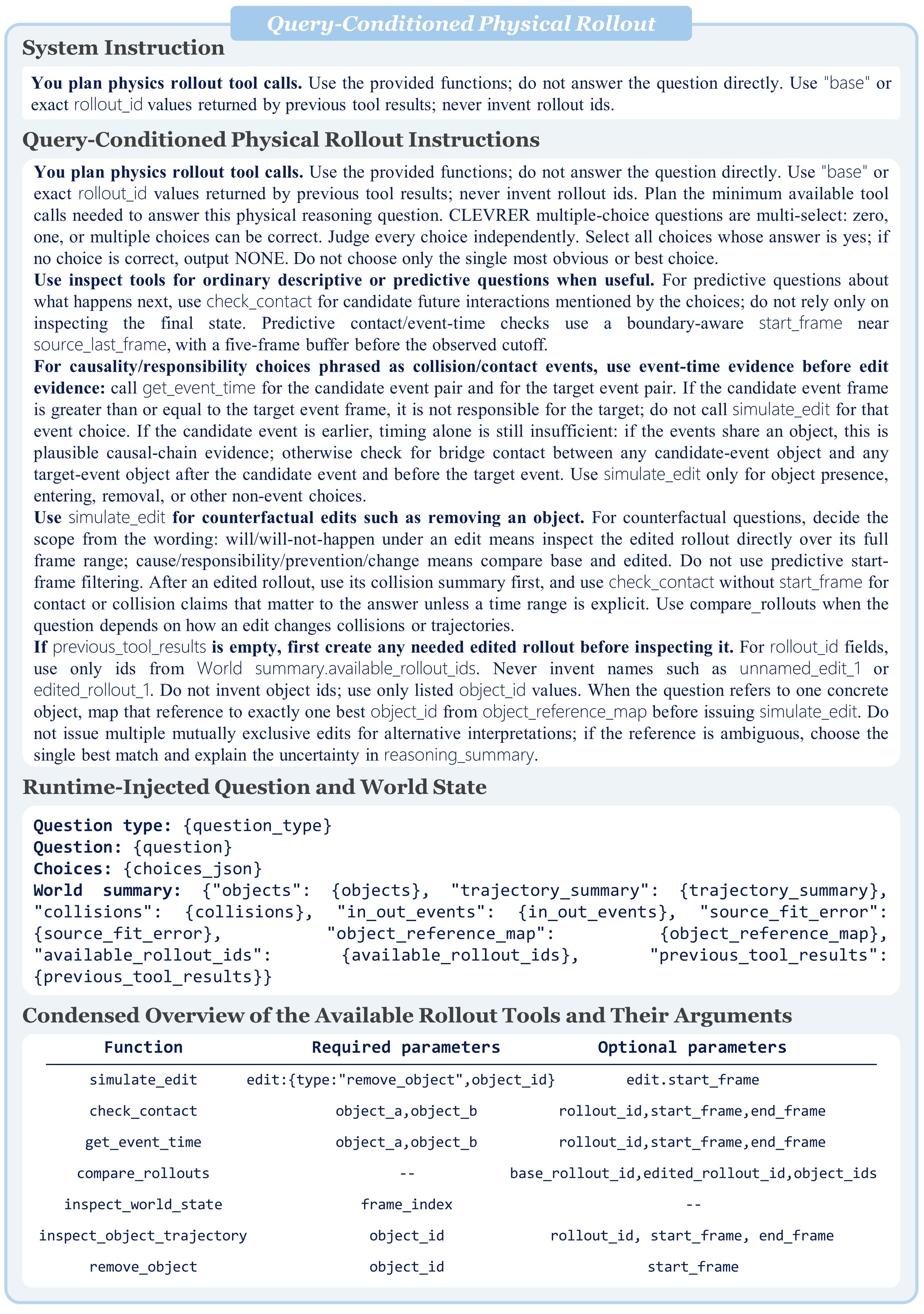}
\caption{Query-conditioned physical rollout. The VLM receives the question and
choices, reconstructed-world summaries, reference grounding, prior tool
results, fixed planning policies, and full callable function schemas. The
embedded table summarizes the available rollout tools and their arguments. All
displayed content is model input; no tool call or rollout result is shown.}
\label{fig:prompt-rollout}
\end{figure*}

\begin{figure*}[p]
\centering
\includegraphics[width=\textwidth,height=0.84\textheight,keepaspectratio]{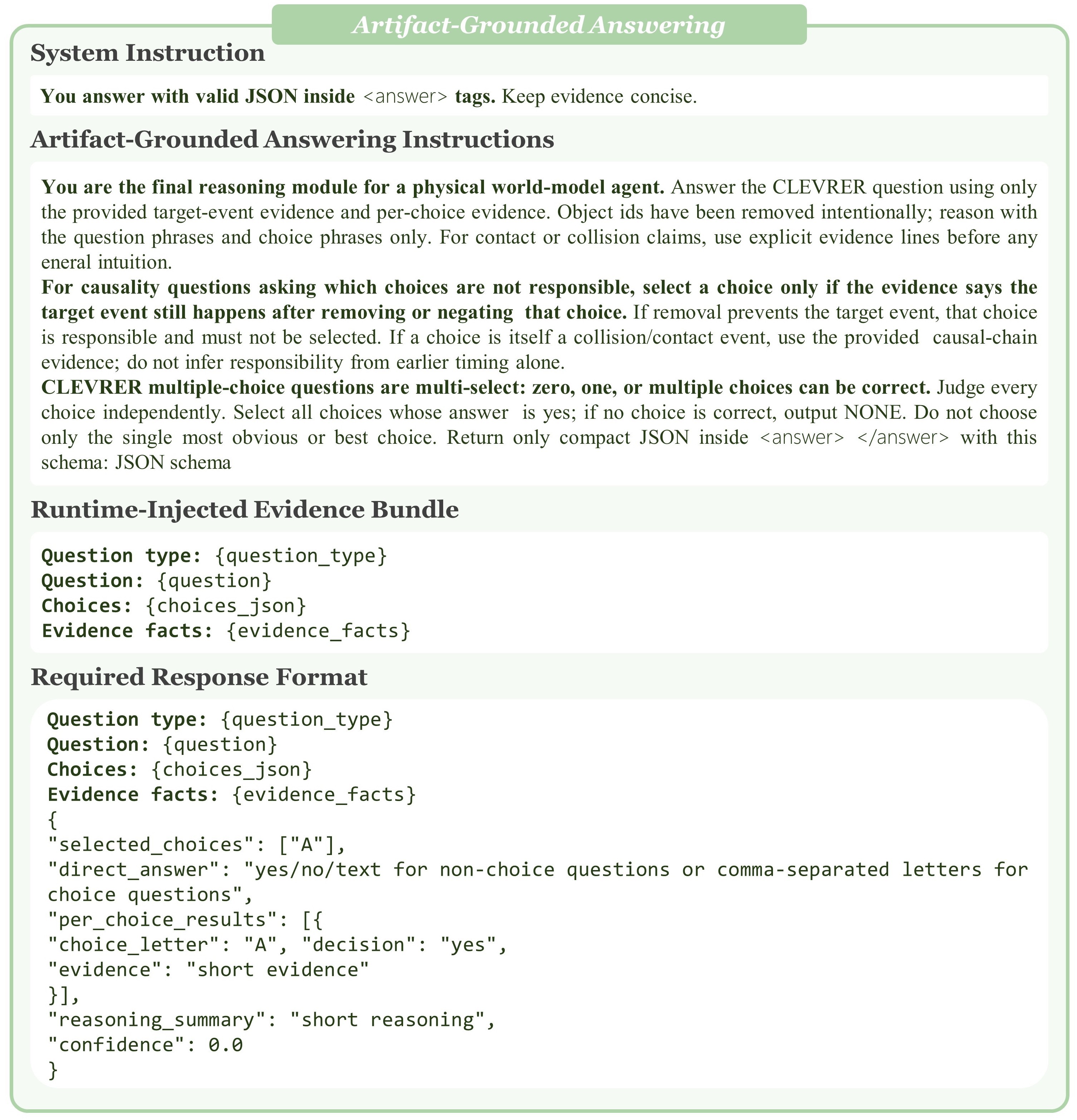}
\caption{Artifact-grounded answering. The answering VLM receives only the
question, choices, and rollout-derived evidence bundle, together with fixed
evidence-use policies and the required JSON response format. The schema is part
of the input instruction; no predicted answer is shown.}
\label{fig:prompt-answering}
\end{figure*}

Figures~\ref{fig:prompt-segmentation}--\ref{fig:prompt-answering} document the
five VLM interfaces used by PhysMind. Each figure separates fixed instructions
from runtime-injected inputs and output constraints. The displayed response
schemas and function definitions specify model interfaces rather than generated
outputs unless stated otherwise.

\begin{table*}[p]
\centering
\small
\begin{tabular}{L{3.45cm}*{12}{C{0.72cm}}}
\toprule
Method
& \multicolumn{2}{c}{Fric. Plat.}
& \multicolumn{2}{c}{Fric. Coll.}
& \multicolumn{2}{c}{Bounce Plat.}
& \multicolumn{2}{c}{Bounce Wall}
& \multicolumn{2}{c}{Mass Coll.}
& \multicolumn{2}{c}{Overall} \\
\cmidrule(lr){2-3}\cmidrule(lr){4-5}\cmidrule(lr){6-7}
\cmidrule(lr){8-9}\cmidrule(lr){10-11}\cmidrule(lr){12-13}
& scene & pair & scene & pair & scene & pair
& scene & pair & scene & pair & scene & pair \\
\midrule
\rowcolor{black!6}
\multicolumn{13}{l}{\textit{Baselines}} \\
Random
& 48.44 & \textbf{21.88}
& 45.31 & \underline{21.88}
& 53.12 & \textbf{27.08}
& 51.04 & \underline{25.00}
& 43.75 & 15.62
& 48.96 & \underline{22.92} \\
Blind Gemini-3-Flash
& 46.88 & 0.00
& 50.00 & 0.00
& 47.92 & 2.08
& 52.08 & 4.17
& 51.56 & 3.12
& 49.74 & 2.08 \\
\midrule
\rowcolor{black!6}
\multicolumn{13}{l}{\textit{Foundation VLMs}} \\
Qwen3-VL-235B-A22B
& 48.44 & 3.12
& 43.75 & 9.38
& \underline{56.25} & \underline{25.00}
& 50.00 & 20.83
& 42.19 & 9.38
& 48.96 & 15.10 \\
Gemini-3-Flash
& \textbf{56.25} & 12.50
& 46.88 & 15.63
& 52.08 & 12.50
& 50.00 & 12.50
& 53.13 & \underline{21.88}
& 51.56 & 14.59 \\
GPT-5.5
& 51.56 & \underline{18.75}
& \textbf{65.62} & \textbf{34.38}
& 55.21 & 18.75
& \underline{58.33} & 20.83
& \underline{60.94} & \underline{21.88}
& \underline{58.07} & 22.40 \\
\midrule
\rowcolor{black!6}
\multicolumn{13}{l}{\textit{Training-Based Methods}} \\
Video-R1
& \underline{54.69} & 9.38
& 53.13 & \underline{21.88}
& 51.04 & 2.08
& 45.83 & 16.67
& 53.13 & 18.75
& 51.04 & 13.02 \\
Chain-of-Frames
& 48.44 & 6.25
& 51.56 & 3.13
& 50.00 & 4.17
& 51.04 & 2.08
& 50.00 & 3.13
& 50.26 & 3.65 \\
\midrule
\rowcolor{black!6}
\multicolumn{13}{l}{\textit{Training-Free Methods}} \\
\textbf{PhysMind}
& \underline{54.69} & 15.63
& \underline{64.06} & \textbf{34.38}
& \textbf{57.29} & \underline{25.00}
& \textbf{59.38} & \textbf{29.17}
& \textbf{64.06} & \textbf{31.25}
& \textbf{59.64} & \textbf{27.09} \\
\bottomrule
\end{tabular}
\caption{Detailed Physion++ object contact prediction accuracy (\%) under the
benchmark-provided terminal red/yellow cue protocol. Scene denotes per-scene
accuracy; pair denotes matched-pair accuracy and requires both trials to be
correct. Fric., Plat., and Coll. abbreviate
friction, platform, and collision. Bold and underline denote the best and
second-best results, respectively.}
\label{tab:physionpp-paired-results}
\end{table*}

\begin{table*}[p]
\centering
\small
\begin{tabular}{L{3.45cm}*{12}{C{0.72cm}}}
\toprule
Method
& \multicolumn{2}{c}{Fric. Plat.}
& \multicolumn{2}{c}{Fric. Coll.}
& \multicolumn{2}{c}{Bounce Plat.}
& \multicolumn{2}{c}{Bounce Wall}
& \multicolumn{2}{c}{Mass Coll.}
& \multicolumn{2}{c}{Overall} \\
\cmidrule(lr){2-3}\cmidrule(lr){4-5}\cmidrule(lr){6-7}
\cmidrule(lr){8-9}\cmidrule(lr){10-11}\cmidrule(lr){12-13}
& scene & pair & scene & pair & scene & pair
& scene & pair & scene & pair & scene & pair \\
\midrule
\rowcolor{black!6}
\multicolumn{13}{l}{\textit{Baselines}} \\
Random
& 48.44 & \underline{21.88}
& 45.31 & \textbf{21.88}
& \textbf{53.13} & \textbf{27.08}
& \underline{51.04} & \textbf{25.00}
& 43.75 & 15.63
& 48.96 & \underline{22.92} \\
Blind Gemini-3-Flash
& 50.00 & 0.00
& \underline{50.00} & 0.00
& 50.00 & 0.00
& 50.00 & 0.00
& \underline{50.00} & 0.00
& 50.00 & 0.00 \\
\midrule
\rowcolor{black!6}
\multicolumn{13}{l}{\textit{Foundation VLMs}} \\
Qwen3-VL-235B-A22B
& 48.44 & 0.00
& \underline{50.00} & \underline{12.50}
& 51.04 & \underline{25.00}
& 45.83 & 14.58
& 43.75 & \underline{18.75}
& 47.92 & 15.10 \\
Gemini-3-Flash
& \textbf{65.63} & \textbf{40.63}
& \underline{50.00} & \underline{12.50}
& 47.92 & \textbf{27.08}
& \textbf{56.25} & \underline{16.67}
& \textbf{53.13} & \textbf{21.88}
& \textbf{54.17} & \textbf{23.44} \\
\midrule
\rowcolor{black!6}
\multicolumn{13}{l}{\textit{Training-Based Methods}} \\
Video-R1
& \underline{57.81} & 15.63
& \textbf{51.56} & 3.13
& \underline{52.08} & 4.17
& 45.83 & 2.08
& \underline{50.00} & 9.38
& \underline{51.04} & 6.25 \\
Chain-of-Frames
& 48.44 & 0.00
& \underline{50.00} & 0.00
& 48.96 & 2.17
& 46.88 & 8.33
& \underline{50.00} & 15.63
& 48.70 & 5.26 \\
\bottomrule
\end{tabular}
\caption{Physion++ baseline accuracy (\%) under persistent red/yellow tracking
cues. Only the cue presentation and prompt differ from
Table~\ref{tab:physionpp-paired-results}. Bold and underline denote the best
and second-best rerun results, respectively.}
\label{tab:physionpp-tracking-cue}
\end{table*}

\begin{table*}[p]
\centering
\small
\begin{tabular}{L{4.35cm}*{8}{C{1.20cm}}}
\toprule
Method
& \multicolumn{2}{c}{Explanatory}
& \multicolumn{2}{c}{Predictive}
& \multicolumn{2}{c}{Counterfactual}
& \multicolumn{2}{c}{Overall} \\
\cmidrule(lr){2-3}\cmidrule(lr){4-5}\cmidrule(lr){6-7}\cmidrule(lr){8-9}
& per ques. & per opt.
& per ques. & per opt.
& per ques. & per opt.
& per ques. & per opt. \\
\midrule
\rowcolor{black!6}
\multicolumn{9}{l}{\textit{Reference}} \\
\textbf{PhysMind}
& \underline{79.44} & \underline{88.31}
& 60.00 & 79.29
& \underline{71.89} & \underline{89.52}
& \underline{73.10} & \underline{87.98} \\
\midrule
\rowcolor{black!6}
\multicolumn{9}{l}{\textit{Component Substitutions}} \\
w/ GeoCalib Up-Direction
& 76.67 & 86.62
& \underline{65.71} & \underline{81.43}
& 69.73 & 88.44
& 71.95 & 86.94 \\
w/o Depth-Aware Filtering
& \textbf{83.89} & \textbf{91.69}
& \textbf{67.14} & \textbf{82.86}
& \textbf{72.97} & \textbf{89.68}
& \textbf{76.55} & \textbf{89.92} \\
w/ GoTrack Pose Tracking
& 78.33 & 88.15
& 64.29 & 79.29
& 68.11 & 86.13
& 71.72 & 86.38 \\
\bottomrule
\end{tabular}
\caption{Additional component ablations on 100 CLEVRER scenes. Per-question
accuracy requires every option of a question to be correct. Bold and underline
denote the best and second-best results in each column, respectively.}
\label{tab:clevrer-additional-ablations}
\end{table*}

\FloatBarrier

\FloatBarrier

\bibliography{references}


\end{document}